# Distributed Data Association in Smart Camera Networks via Dual Decomposition

**Wan Jiuqing**

Department of Automation

Beijing University of Aeronautics and Astronautics

wanjiuqing@buaa.edu.cn

**Nie Yuting**

Department of Automation

Beijing University of Aeronautics and Astronautics

nieyuting@buaa.edu.cn

**Liu Li**

Department of Automation

Beijing University of Aeronautics and Astronautics

liulizi123@sina.com

**Abstract:** One of the fundamental requirements for visual surveillance using smart camera networks is the correct association of each person's observations generated on different cameras. Recently, distributed data association that involves only local information processing on each camera node and mutual information exchanging between neighboring cameras has attracted many research interests due to its superiority in large scale applications. In this paper, we formulate the problem of data association in smart camera networks as an Integer Programming problem by introducing a set of linking variables, and propose two distributed algorithms, namely L-DD and Q-DD, to solve the IP problem using dual decomposition technique. In our algorithms, the original IP problem is decomposed into several subproblems, which can be solved locally and efficiently on each smart camera, and then different subproblems reach consensus on their solutions in a rigorous way by adjusting their parameters iteratively based on projected subgradient optimization. The proposed methods are simple and flexible, in that (i) we can incorporate any feature extraction and matching technique into our framework to measure the similarity between two observations, which is used to define the cost of each link, and (ii) we can decompose the original problem in any way as long as the resulting subproblem can be solved independently on individual camera. We show the competitiveness of our methods in both accuracy and speed by theoretical analysis and experimental comparison with state of the art algorithms on two real datasets collected by camera networks in our campus garden and office building.

1. Introduction

Many research interests arose in recent years regarding wide-area surveillance using networked smart cameras with non-overlapping Field of Views (FOVs). The camera nodes constituting the networks are smart, that is, they are not only able to collect video data, but also capable of local computation and mutual communication. They usually work cooperatively to discover or understand the behavior of some objects of interest, such as pedestrians or vehicles, moving in the monitored region. One of the fundamental prerequisites for achieving these goals is the correct reconstruction of camera-to-camera trajectory of each object, or equivalently, grouping observations originated from the same object into a single track, which may be generated by different cameras at different time instants. This problem is often referred to as data association in camera networks [1,12], trajectory recovery [13], or camera-to-camera tracking [14-18]. In this paper, we assume that the detection and tracking problem within a single camera view has been solved, and we call the collection of quantities summarizing the features of tracked object as a virtual "observation", see Fig.2 as an example.

However, data association in camera networks is a rather challenging task. First, the visual appearance features may be indistinctive, e.g. persons are difficult to distinguish from one another if they wear similar clothes. In addition, as the cameras are mounted at different sites with various view angles and lighting conditions, the appearance of an object may undergo large variations across disjoint camera views, resulting in different objects appearing more alike than that of the same object. This makes the performance of the methods based solely on appearance cues [2-6] far from satisfactory. Second, in the methods for multi-targets tracking or data association in single [7-8] or overlapping views [29-31], some motion model is usually assumed to be valid which imposes strong constraint on the object's state, and can be used to distinguish different objects even they look very similar. Unfortunately, because large unobserved areas exist between disjoint camera views, this kind of assumption no longer holds true for our problem. Instead, some spatio-temporal cues, such as when and where the observation was made, could be exploited to improve the data association accuracy if some knowledge about the layout of the monitored region were available [9]. For example, the distance of two camera sites imposes a constraint on the minimum traveling time of a person moving between the two cameras. However, as the traveling time between disjoint cameras varies greatly from person to person, and there may be unpredictable pauses in blind regions, it is hard to impose accurate spatio-temporal constraints. Third, computationally, data association is intrinsically a combinatorial problem. Unless the special structure of some specific formulation of the problem were exploited, such as in [10,21], for data association problem of moderate or large scale only approximate solution can be obtained with reasonable computational efforts. The limitation in computation plus the weakness of appearance and spatio-temporal cues render accurate data association in camera networks very difficult. Finally, for large scale camera networks, it is unrealistic to transmit video data collected by all cameras to a central server for processing due to the limitation in communication bandwidth. Although smart camera can analyze local video and transmit only extracted features, the central server will become overwhelmed quickly when the number of observations increases because of the combinatorial nature of data association problem. Thus, distributed algorithm, which involves only local computation on each camera and information exchanging between neighboring cameras, is preferred for the sake of scalability of system, robustness against single point of failure, and efficiency in use of bandwidth [11]. However, finding global optimal data association solution based solely on local computation and communication is far from a trivial matter.

1.1 Related works

*A. Centralized data association*

The problem of data association in camera networks can be treated in a Bayesian framework [1,12], in which each observation is assigned with a labeling variable, and the posterior distribution of each labeling variables are inferred based on all appearance and spatio-temporal evidence made in the whole networks. The resulting marginal distribution of labeling variable contains the complete knowledge about which object the observation came from and the trajectory of moving object can be recovered directly using these knowledge. However, doing inference in the joint labeling space is usually intractable and the authors have to resort to some assumed independence structure and approximate algorithms such as Assumed Density Filter [19].

Some authors try to solve the data association problem by optimally partitioning the set of observations collected by camera networks into several disjoint subsets according to assumed appearance and spatio-temporal models, such that the observations in each subset are believed to come from a single object. The difficulty caused by the exponential explosion of the partition space is tackled by making appropriate independence assumptions and leveraging efficient optimization algorithms, such as Markov Chain Monte Carlo [13], Multiple Hypothesis Tracking [17,18], or some heuristic optimization scheme [20]. It is noticeable that in some works the data association problem is mapped into a problem of finding disjoint flow paths in a cost-flow network, and each resulting path corresponds to the trajectory of a single object [10,16,21]. The nodes of the network correspond to observations, and the costs of edge are defined by the similarity between the connected observations. The global optimal flow paths that minimize the total cost can be found efficiently by exploiting the special structure of the problem and using algorithms such as Min-cost Flow [21,16], or K-shortest Path optimization [10]. The main limitation of the network flow methods is that they can only model linear constraints, i.e. the similarity between adjacent observations in a path. A new graph model was proposed recently [23], which is capable of capturing higher order information, such as the smoothness of the trajectory. However, this model leads to a problem that can no longer be solved by Min-cost Flow algorithm. The authors proposed an iterative algorithm based on Lagrangian relaxation, transforming the original problem into a series of problems that are solvable by min cost flow.

However, the above works are all centralized in that the observations generated on all cameras in the networks need to be transmitted to a central server where some data association algorithm is running, which is undesirable for large scale applications due to the limitation of camera energy and communication bandwidth.

*B. Distributed data association*

Recently, distributed data association in camera networks, which involve only local information processing in each camera node and information exchanging between neighboring cameras while being able to achieve the same or similar association results as their centralized counterparts, have attracted many research interests.

Considering the appearance observations made in the networks as i.i.d. samples drawn from a mixture model, and treating the labeling variables as missing data, various kinds of distributed EM algorithms can be used for appearance based distributed data association [24-25]. However, theses algorithms always perform poorly when observing conditions vary largely across camera views, as they are based on the assumption that the appearance observations of a single object follows a unimodal, usually Gaussian, distribution. To improve the performance, exploiting spatio-temporal information is necessary. Unfortunately, unlike the case of traditional wireless sensor networks [26-28] or camera networks with overlapping views [29-31], where the dependence of involved variables in spatial dimension (intra-scan dependence) and temporal dimension (inter-scan dependence) can be modeled separately, the spatial and temporal evidence made in non-overlapping camera networks are tightly coupled. This precludes the use of most existing distributed inference or optimization algorithms for traditional WSN and overlapping camera networks. To overcome these difficulties, in [12] the authors proposed a spatio-temporal tree to model the dependence structure of the involved variables and use belief propagation algorithm for calculate the posterior distribution of each labeling variables in a distributed manner. The main limitation of [12] is that it requires the knowledge of the number of objects under tracking, which is usually unavailable in practice. In a recent work [32], the authors proposed a distributed Bayesian framework in which both the sampling space and the posterior distribution of labeling variables are inferred online, based on knowledge propagation between neighboring cameras. The number of objects can be determined automatically from observation data.

Besides distributed inference methods, there are also some works in which the problem of data association is solved through distributed optimization. In the approach of [33], each camera independently estimate local paths of objects in its neighborhood based on observations sent from its neighboring cameras and local spatio-temporal model. The conflicts on locally estimated paths among cameras are resolved by a voting algorithm, and the agreed local paths are finally combined into global paths. The main limitation of [33] is that the non-overlap constraint on paths of different objects is not expressed explicitly during optimization and the error caused by violation of this constraint cannot be necessarily recovered by voting process.

## 1.2 Our contributions

In this paper, we consider the problem of data association in camera networks as recovering the camera-to-camera trajectory of each person by linking observations originated from him or her, and propose a powerful and flexible framework based on Dual Decomposition [34-35] to find the optimal linking configuration for the set of observations collected by the whole camera networks in a distributed manner. The main contributions include:

(i) We formulate the optimal linking problem as a constrained Integer Programming problem. Specifically, we assign a binary linking variable to each possible link between observation pairs to indicate whether it takes effect, and the collection of all linking variables constitutes the optimization variable of the IP problem. The linking variables do not take value arbitrarily. They must satisfy the constraint that each observation should not be included in more than one trajectories. Then we define two kinds of energy function to measure the likelihood of different linking configurations. The first is a linear energy function which is calculated by adding up the similarities between adjacent observations of each trajectory; the second is a quadratic energy function which is capable of evaluating higher-order similarity of three-observation tracklets. We calculate the similarity measure using very simple appearance and spatio-temporal models, but more advanced feature extraction and matching techniques [2-6] can be applied easily in our framework.

(ii) The above IP problem is very difficult to solve because the searching space is extremely large. In this paper, we apply the dual decomposition technique to our problem, which decomposes a large problem into several solvable subproblems and enforce consistency of solutions on common variables between different subproblems by adjusting parameters of each subproblems in a rigorous way. Specifically, we propose two algorithms: L-DD and Q-DD. L-DD is used for solving the linear energy minimization problem. L-DD decompose the original IP problem into $2K$ (two for each camera, $K$ is the number of cameras in the network) linear assignment subproblems, which are solved independently on each smart camera by using Hungarian algorithm. Q-DD is used for solving the quadratic energy minimization problem. Q-DD decomposes the IP problem into $N$ (one for each observation, $N$ is the number of observations) subproblems, which are solved on each camera by direct searching. In L-DD and Q-DD, the parameters of each subproblems are updated at each iteration by using the projected subgradient method that maximizes the lower bound of energy function and enforces the consistency on solutions between overlapping subproblems. Both L-DD and Q-DD can be implemented in a distributed manner, that is, it involves only local operations on each camera and mutual information exchanges between neighboring cameras. It is also worthy to note that the DD framework is very flexible in that we can design new algorithm by considering other kinds of decomposition as long as the resulting subproblems can be solved on each camera efficiently.

(iii) We analyze the proposed algorithms theoretically and experimentally. For L-DD, we prove that the computed energy lower bound is guaranteed to reach the optimal value of the energy function. In other words, L-DD is guaranteed to find the global optimal solution for the linear energy model. For Q-DD, we prove that the optimal lower bound of the quadratic energy function provided by our method is the same as that provided by a recently proposed centralized data association algorithm [23] which is based on Lagragian relaxation of the original problem. In addition, we show the effectiveness of our algorithms by applying them to two real datasets and comparing with the state-of-the-art methods. The two datasets are collected by camera networks consisting of 10 cameras mounted on our campus garden and office building, respectively. Our algorithms show superiority in both accuracy measured by precision, recall and F-measure, and speed measured by running time of Matlab codes on a PC.

The rest of this paper is organized as follows: In Sec.2 we describe the data association problem formally and transform it into integer programming problem by introducing a set of binary linking variables. We define two kinds of energy model and discuss the calculation of model parameters. In Sec.3 we introduce the Dual Decomposition framework and propose two algorithms, L-DD and Q-DD, for minimizing the linear and quadratic energy functions, respectively. We also discuss the property of the lower bounds given by our algorithms theoretically. In Sec.4 we report the experimental results of our methods and compare it with some the-state-of-the-are methods, showing the superiority of our algorithms. Conclusion and further research perspective are given in Sec.5.

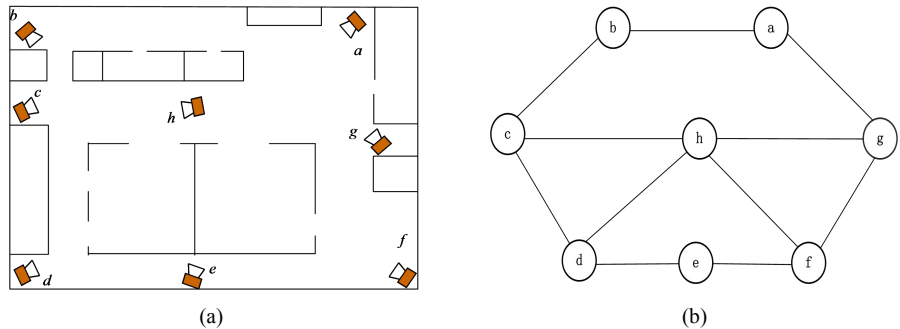

(a)     (b)

Fig.1 Smart camera networks and its topology

2.1 Problem description

Suppose that multiple objects are moving in a large area monitored by a set of smart cameras with non-overlapping field of views, as shown in Fig.1(a). Here the number of moving objects is unfixed in time and unknown to us. We assume that each camera node has limited resource for computation, storage and communication, and synchronized internal clocks that allow the nodes to share a common notion of time. The camera networks can be represented as a graph $\mathcal{G} = (\mathcal{V}, \mathcal{E})$, which is often called camera networks topology or activity topology, see Fig.1(b). Each camera corresponds to a node $v \in \mathcal{V}$ in the graph, and two nodes are linked by an edge $e \in \mathcal{E}$ if it is assumed that object can move from one camera to another without being observed by other cameras. We will call this assumption as topological assumption hereafter. In this paper, camera node $v$ is called neighbor of node $u$ if $(u,v) \in \mathcal{E}$.

Now we explain the meaning of the word "observation" in this paper. When an object is passing, a clip of video is collected by the observing camera, as shown in Fig.2(a). In this paper, we assume that the collected video data is summarized into a single virtual observation $y_i$, using some visual tracking and feature extraction algorithms running on the camera. We further assume that each observation $y_i$ is originated from a single object. If multiple objects present in a camera's FOV simultaneously, we track them with some multi-object tracking algorithm and extract one observation for each object. The index $i$ implies that $y_i$ is the $i$ th observation generated by the whole networks in time order. Each observation $y_i = (o_i, d_i)$ consists of two parts: the appearance measurement $o_i$, such as the size, texture, color distribution, or biometric features of the object; and the spatio-temporal measurement $d_i$, such as the capturing time or the moving direction of the object in the camera's FOV. See Fig.2 (b) and (c) for an example.

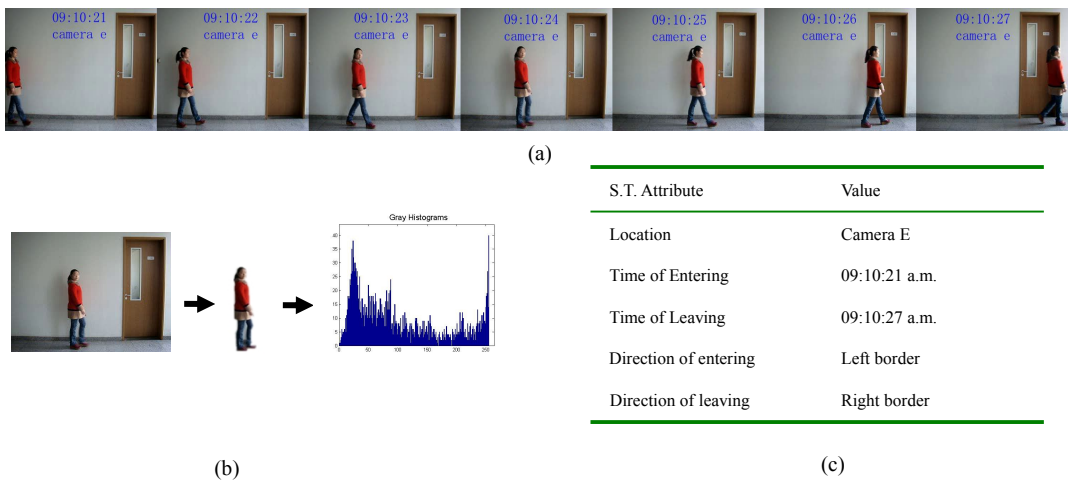

Fig.2 Observations made by camera node. (a) Video frames collected by camera when an object is passing by. (b) Appearance observation: the color histogram of the object region segmented from frames. (c) Spatio-temporal observation: the time and direction of the object's entering in or leaving from the camera's FOV.

After a certain period of time, a set of observations $Y = \{y_i\}$, $i \in \{1:N\}$ are generated by the networks. The task of data association is to find a partition of $Y$ into several subsets (or tracks) such that the observations in each subset are believed to come from a single object. A valid partition is a division of $Y$ into non-empty subsets which are both collectively exhaustive and mutually exclusive with respect to $Y$. It is preferable in large scale camera networks that the data association is performed in a distributed manner, that is, we do not transmit all observations to a central server, but find the optimal partition based only on local calculation on each camera and information exchange between neighboring cameras.

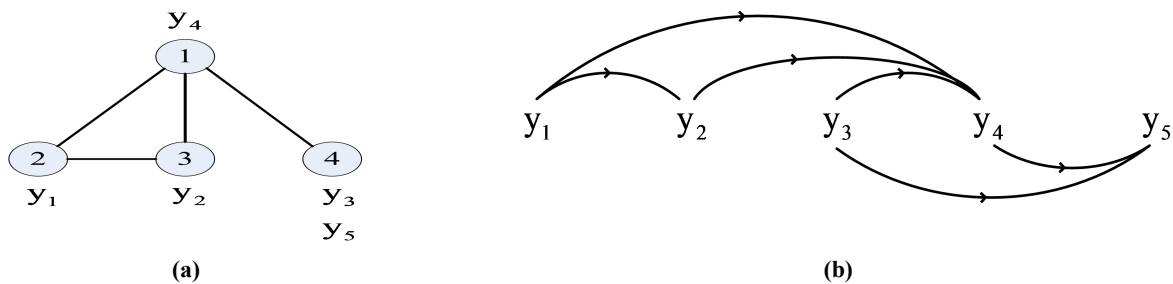

Fig.3 Illustrating example. (a) Camera networks topology. (b) Candidate links between observations. There is a link between $y_1$ and $y_4$, as an object may move from camera 2 to 1, generating observations $y_1$ and $y_4$ consecutively. No link exists between $y_1$ and $y_3$, as according to the networks topology, object cannot move directly from camera 2 to 4 without being observed by other cameras in the networks.

## 2.2 Energy minimization

In this subsection we formulate the problem of data association in camera networks as an energy minimization problem by introducing a set of linking variables. See the simple example in Fig.3. Fig.3(a) shows the topology of a small camera networks. Each node corresponds to a camera, associated with observations generated on it. The observations are indexed in time order. Fig.3(b) shows the candidate links between observations according to the networks topology, which are represented by directional arcs pointing from older observations to newer ones. Two observations are linked if they are believed to be adjacent to each other in the recovered track of a single object. Note that camera networks topology impose strong constraints on the set of candidate links. Indeed, in the recovered track, the immediate predecessor or successor of observation made on camera $u$ can only be selected from observations made on $u$'s neighbors. In case of missing detection, the possible predecessor or successor should be selected in an enlarged neighborhood of $u$. We refer the reader to [36] for the treatment in detail, but we will concentrate on non-missing detection case in this paper. The set of candidate links also contains arcs pointing from a virtual "source" observation $y_0$ to ordinary observations $y_i$, implying that $y_i$ is the head of a track; and arcs pointing from $y_i$ to a virtual "sink" observation $y_\infty$, implying that $y_i$ is the tail of a track. We do not show these virtual links in Fig.3(b) in sake of clarity. In this paper, we call the observations generated by cameras in the networks as ordinary observation, and the links connecting two ordinary observations as ordinary links. We call the links connecting an ordinary observation and a virtual observation, i.e., $y_0$ or $y_\infty$, as virtual links.

We denote the set of incoming links to ordinary observation $y_i$ as $A^-(y_i)$, connecting $y_i$ with its possible predecessors; and the set of outgoing links from $y_i$ as $A^+(y_i)$, connecting $y_i$ with its possible successors. We define $A(y_i) = A^-(y_i) \cup A^+(y_i)$. The set of all candidate links is $A = \bigcup_{i=1}^{N} A(y_i)$. We assign a binary variable $x_q \in \{0,1\}$, called linking variable, to each link in the set $A$, allowing the corresponding link to be active ($x_q = 1$) or inactive ($x_q = 0$). A partition of observation set $Y$ into tracks can be represented as a binary valued vector $\mathbf{x} = \{x_q\}, q \in A$, such that $x_q = 1$ for all links connecting adjacent observations in tracks, and $x_q = 0$ for other links. The vector $\mathbf{x}$ does not take values arbitrarily. In any track of a valid partition, the immediate predecessor or successor of each ordinary observation should be unique. This is called *uniqueness constraint* in this paper. Enforcing these constraints on linking configuration is equivalent to requiring that vector $\mathbf{x}$ lies in the set

$$X = \left\{ \mathbf{x} = \{x_q\} \middle| \begin{array}{l} \text{binary: } x_q \in \{0,1\}, \quad \forall q \in A \\ \text{uniqueness: } \sum_{p \in A^-(y_i)} x_p = \sum_{q \in A^+(y_i)} x_q = 1, \quad \forall i \in \{1:N\} \end{array} \right\} \quad (1)$$

It is easy to see that if the topological assumption (see Fig.1(b)) of the camera networks hold true, there is a one-to-one correspondence between the set $X$ and the set of valid partitions of $Y$.

Next, we define a parameterized energy function $E(\mathbf{x}|\boldsymbol{\theta})$ to evaluate the quality of recovered tracks defined by the vector $\mathbf{x}$. This will allows us to find the optimal partition of $Y$ by minimizing the energy with respect to $\mathbf{x}$ over the constraint set $X$. In this paper, we consider two kinds of energy functions, leading to two different energy minimization problems. The first energy minimization problem considered in this paper is

$$\begin{cases} \min_{\mathbf{x}} E_1(\mathbf{x}|\boldsymbol{\theta}) = \sum_{q \in A} \theta_q x_q \\ \text{s.t.} \quad \mathbf{x} \in X \end{cases} \quad (2)$$

where the energy is a linear function of $\mathbf{x}$. The parameters $\theta_q$ represents the disparity between observations linked by $q$.

Thus, minimizing energy $E_1(\mathbf{x}|\boldsymbol{\theta})$ favors grouping "similar" observations into a single track. We will discuss the calculation of $\theta_q$ in Sec.2.3. Problem (2) is an Integer Linear Programming (ILP) problem with special structure. Specifically, the coefficient matrix of constraint set $\mathbf{X}$ can be viewed as the incidence matrix of a bipartite graph and satisfies the Totally Uni-modular (TUM) property [37]. Indeed, problem (2) is a typical linear assignment problem, which can be solved exactly in polynomial time by Hungarian algorithm. It can also be formulated as a network flow problem and solved efficiently using Min-Cost Flow (MCF) algorithm [21] or K-Shortest Path (KSP) algorithm [10]. However, the above algorithms are all centralized in nature. We will show how to solve the problem exactly in a distributed manner using Dual Decomposition in Sec.3.

The modeling power of $E_1(\mathbf{x}|\boldsymbol{\theta})$ is limited in that the quality of recovered tracks are evaluated based only on similarities between adjacent observations in tracks. Encoding higher order constraints of the tracks in the energy function is, hopefully, able to improve the quality of solution. We note that this has motivated many recent works in multi-target tracking based on single camera such as [23,8]. However, little work has been proposed regarding the use of higher order model for tracking in wide area based on non-overlapping camera networks. In fact, higher order model can encodes our prior knowledge about the movement of object that cannot be effectively captured before. In this paper, we consider the predecessor-successor pair (p-s pair in short) of each ordinary observation. The set of all p-s pairs of an ordinary observation $y_i$ is denoted by $B(y_i) = \{(p,q) | p \in A^-(y_i), q \in A^+(y_i)\}$, and the set of all possible p-s pairs is $B = \bigcup_{i=1}^{N} B(y_i)$. We assign a binary variable $x_{pq} = x_p x_q$ to each p-s pair in the set $B$, and denote $\bar{\mathbf{x}} = \{\{x_q\}, \{x_{pq}\}\}$. The second energy minimization problem considered in this paper is

$$\begin{cases} \min_{\bar{\mathbf{x}}} E_2(\bar{\mathbf{x}}|\bar{\boldsymbol{\theta}}) = \sum_{q \in A} \theta_q x_q + \sum_{pq} \theta_{pq} x_{pq} \\ \text{s.t.} \quad \bar{\mathbf{x}} \in \bar{\mathbf{X}} \end{cases} \quad (3)$$

where $\bar{\boldsymbol{\theta}} = \{\{\theta_q\}, \{\theta_{pq}\}\}$. The parameter $\theta_{pq}$ represents some quality measure of the tracklet consisting of the three adjacent observations connected by links $p$ and $q$. In this paper, we use $\theta_{pq}$ to describe the disparity between the predecessor and the successor of a single observation. The calculation of $\theta_{pq}$ will be discussed in Sec.2.3. The constraint set $\bar{\mathbf{X}}$ is defined as

$$\bar{\mathbf{X}} = \left\{ \bar{\mathbf{x}} = \{x_q, x_{pq}\} \middle| \begin{array}{l} x_q \in \{0,1\}, \forall q \in A, \quad x_{pq} \in \{0,1\}, \forall (p,q) \in B \\ \sum_{p \in A^-(y_i)} x_p = \sum_{q \in A^+(y_i)} x_q = 1, \quad \forall i \in [1:N] \\ x_{pq} \leq x_p, x_{pq} \leq x_q, x_{pq} \geq x_p + x_q - 1, \quad \forall (p,q) \in B \end{array} \right\} \quad (4)$$

Problem (3) is also an ILP problem, but the coefficient matrix of $\bar{\mathbf{X}}$ is not TUM in general. Thus, standard techniques such as Linear Programming (LP) relaxation cannot be guaranteed to find the optimal solution of (3). Indeed, with manageable computational effort, only approximate solutions can be obtained. In Sec.3, we will show how to design approximate algorithm using Dual Decomposition technique to solve (3) in a distributed manner.

2.3 Model parameters

In this subsection, we discuss how to calculate the parameters $\theta$ of the energy functions. Recall that $\theta_q$ evaluates the quality of the tracklet consisting of two observations $(y_i, y_j)$ linked by $q$, such that $q \in A^+(y_i) \cap A^-(y_j)$, and $\theta_{pq}$ evaluates the quality of the tracklet consisting of three observations $(y_i, y_k, y_j)$ linked by $p$ and $q$, such that $p \in A^+(y_i) \cap A^-(y_k)$ and $q \in A^+(y_k) \cap A^-(y_j)$. We define $\theta_\varepsilon = -\ln \phi_\varepsilon$, where $\varepsilon \in \{q, pq\}$, and $\phi_\varepsilon$ is a similarity measure between the linked observations. As each observation includes an appearance part and a spatio-temporal part, we factorize $\phi_\varepsilon$ as $\phi_\varepsilon = \phi_\varepsilon^{ap} \cdot \phi_\varepsilon^{st}$, where $\phi_\varepsilon^{ap}$ represents the appearance similarity and $\phi_\varepsilon^{st}$ the spatio-temporal similarity.

**Calculation of** $\phi_q^{ap}(o_i, o_j)$. In this paper, we take the normalized histogram of RGB brightness values computed over lower and upper part of the object region as appearance measurement. We evaluate the appearance similarity in two ways. First, we directly compute the Bhattacharya distance between $o_i$ and $o_j$, and defeine $\phi_q^{ap}$ as

$$\phi_q^{ap}(o_i, o_j) = 1 - Bhatt(o_i, o_j) \tag{5}$$

Second, follows [38], we map the appearance on one camera to another by Cumulative Brightness Transfer Functions (CBTF) established for each pair of camera sites using training data, and compute the Bhattacharya distance between the transferred appearance measurements. The similarity $\phi_q^{ap}$ is calculated as

$$\phi_q^{ap} = 1 - \tfrac{1}{2} Bhatt(o_i, \hat{o}_j) - \tfrac{1}{2} Bhatt(\hat{o}_i, o_j) \tag{6}$$

where $\hat{o}$ represents mapped appearance measurements using CBTF. In general, Eq.(6) can estimate appearance similarity more properly when object appearance undergoes significant changes across different camera sites. More advanced feature representation and distance computation techniques for appearance or biometric based object re-identification [2-6] can be exploited here to calculate $\lambda_{ap}$, but we leave this for further study.

**Calculation of** $\phi_q^{st}(d_i, d_j)$. In this paper, we use the camera site $l$, the time entering FOV $t^{en}$, time leaving FOV $t^{le}$, direction of entering in image $e^{en}$ and direction of leaving in image $e^{le}$, as spatio-temporal measurements. We factorize $\phi_q^{st}$ as

$$\phi_q^{st}(d_i, d_j) = p_t(t_j^{en} | t_i^{le}) p_e(l_j, e_j^{en} | l_i, e_i^{le}) \tag{7}$$

The first factor is the traveling time distribution

$$p_t(t_i^{le}, t_j^{en}) = \begin{cases} 0, & \text{other} \\ 1, & \Delta_{i,j}^{\min} \leq t_j^{en} - t_i^{le} \leq \Delta_{i,j}^{\max} \end{cases} \tag{8}$$

where $\Delta_{i,j}^{\min}$ and $\Delta_{i,j}^{\max}$ are the minimum and maximum traveling time between camera sites $l_i$ and $l_j$, respectively. Note that in previous works truncated Gaussian distribution was used as traveling time model [1,12,32]. Our traveling time constraint is much weaker than Gaussian, but it is more suitable when person walks in abnormal speed, for example, pauses randomly during walking. The second factor in Eq.(7) is a discrete distribution specifying the probability of an object arriving camera site $l_j$ in direction $e_j^{en}$ when departing from camera site $l_i$ in direction $e_i^{le}$. The parameters of the spatio-temporal model are learned from training data or specified according to prior knowledge.

**Calculation of** $\phi_{pq}^{ap}(o_i, o_j)$. $\phi_{pq}^{ap}$ measures the similarity between the appearance parts of the first and the third observations of the tracklet $y_i - y_k - y_j$, which is defined by links $p$ and $q$. The calculation of $\phi_{pq}^{ap}$ is similar to $\phi_q^{ap}$.

**Calculation of** $\phi_{pq}^{st}(l_i, l_k, l_j)$. The spatio-temporal quality of the tracklet $y_i - y_k - y_j$ is defined as the fittingness of the sequence $l_i - l_k - l_j$ to a pre-leaned 2-order Markov chain model

$$\phi_{pq}^{st}(l_i, l_k, l_j) = p(l_j | l_k, l_i) \tag{9}$$

The 2-order Markov chain can express more complex causality in the movement of object, e.g., it is very likely that an object at location $B$ will move to $C$ if it arrived $B$ from $A$. The model parameters can be leaned from training data or specified according to prior knowledge.

## 3. Algorithms

### 3.1 Dual decomposition

Recently, Dual Decomposition (DD) technique has been used in many works to solve the discrete optimization problems, showing its extreme generality, flexibility and effectiveness [34-35,39]. In this paper, we will exploit DD technique to solve the energy minimization problems proposed in Sec.2. We consider the following IP problem, which includes problem (2) and (3) as special cases

$$\begin{cases} \min_{\mathbf{x}} E(\mathbf{x}|\boldsymbol{\theta}) = \langle \mathbf{x}, \boldsymbol{\theta} \rangle \\ \text{s.t.} \quad \mathbf{x} \in X \end{cases} \tag{10}$$

In (10), $\mathbf{x} = \{x_q\}$ is a binary-valued vector of linking variables, the cost function $E$ is linear with respect to $\mathbf{x}$ and parameterized by $\boldsymbol{\theta}$, and the constraint $X$ is a subset of $\{0,1\}^n$, where $n$ is the number of elements in $\mathbf{x}$. The basic idea of DD is to decompose the original hard optimization problem into a set of easier subproblems, and find the solution to original problem through the cooperation between subproblem solvers. Let $\Upsilon$ be a collection of (overlapping) subsets of links, ensuring that each link $q$ is included in at least one subset $\sigma \in \Upsilon$. For each $\sigma$, we define a vector of linking variables $\mathbf{x}^\sigma = \{x_q^\sigma\}_{a \in \sigma}$, a corresponding vector of parameters $\boldsymbol{\theta}^\sigma$, and a constraint $X^\sigma$ on $\mathbf{x}^\sigma$. Then the $\sigma$ th subproblem is

$$\text{SP}^\sigma : \min_{\mathbf{x}^\sigma \in X^\sigma} \langle \mathbf{x}^\sigma, \boldsymbol{\theta}^\sigma \rangle \tag{11}$$

Here we assume that the global optimizer of $\text{SP}^\sigma$ can be found with reasonable computational efforts. If the following conditions are satisfied, we say that the set of subproblems $\{\text{SP}^\sigma\}_{\sigma \in \Upsilon}$ provides a decomposition of the original problem (10)

$$\langle \mathbf{x}, \boldsymbol{\theta} \rangle = \sum_\sigma \langle \mathbf{x}^\sigma, \boldsymbol{\theta}^\sigma \rangle \tag{12}$$

$$\{\{\mathbf{x}^\sigma\}, \mathbf{x} | \mathbf{x}^\sigma \in X^\sigma, \mathbf{x}^\sigma = \mathbf{x}_{|\sigma}, \forall \sigma\} = \{\{\mathbf{x}^\sigma\}, \mathbf{x} | \mathbf{x} \in X, \mathbf{x}^\sigma = \mathbf{x}_{|\sigma}, \forall \sigma\} \tag{13}$$

where $\mathbf{x}_{|\sigma}$ represents the subvector of $\mathbf{x}$ corresponding to the same subset of links as $\mathbf{x}^\sigma$. Then we get an equivalent form of problem (10)

$$\begin{cases} \min_{\{\mathbf{x}^\sigma\}, \mathbf{x}} \sum_\sigma \langle \mathbf{x}^\sigma, \boldsymbol{\theta}^\sigma \rangle \\ \text{s.t.} \quad \mathbf{x}^\sigma \in X^\sigma, \mathbf{x}^\sigma = \mathbf{x}_{|\sigma}, \forall \sigma \end{cases} \tag{14}$$

The constraint $\mathbf{x}^\sigma = \mathbf{x}_{|\sigma}$ requires the overlapping vectors $\mathbf{x}^\sigma$ to be consistent on their values of common links. We relax this constraint by introducing Lagrange multipliers $\{\boldsymbol{\lambda}^\sigma\}$ and write the Lagrange dual function as

$$g(\{\boldsymbol{\lambda}^\sigma\}) = \min_{\{\mathbf{x}^\sigma \in X^\sigma\}, \mathbf{x}} \sum_\sigma \langle \mathbf{x}^\sigma, \boldsymbol{\theta}^\sigma \rangle + \sum_\sigma \langle \boldsymbol{\lambda}^\sigma, \mathbf{x}^\sigma - \mathbf{x}_{|\sigma} \rangle \tag{15}$$

which yields a lower bound on the optimal value of problem (14). Using dual feasibility, $\mathbf{x}$ can be eliminated from (15) by imposing a constraint on $\{\boldsymbol{\lambda}^\sigma\}$

$$\Lambda = \left\{ \{\boldsymbol{\lambda}^\sigma\} \middle| \sum_{\sigma \in \Upsilon(q)} \lambda_q^\sigma = 0, \quad \forall q \right\} \tag{16}$$

where $\Upsilon(q)$ represents the collection of subsets containing link $q$. Now we get the dual problem of (14) as

$$\begin{cases} \max_{\{\boldsymbol{\lambda}^\sigma\}} g(\{\boldsymbol{\lambda}^\sigma\}) = \sum_\sigma g^\sigma(\boldsymbol{\lambda}^\sigma) \\ \text{s.t.} \quad \{\boldsymbol{\lambda}^\sigma\} \in \Lambda \end{cases} \tag{17}$$

where

$$g^\sigma(\boldsymbol{\lambda}^\sigma) = \min_{\mathbf{x}^\sigma \in X^\sigma} \langle \mathbf{x}^\sigma, \boldsymbol{\theta}^\sigma + \boldsymbol{\lambda}^\sigma \rangle \tag{18}$$

is the optimal value of subproblem $\text{SP}^\sigma$ whose parameter is $\boldsymbol{\theta}^\sigma + \boldsymbol{\lambda}^\sigma$. It is noticeable that the cost function in (17) is concave as it is the sum of pointwise minimums of sets of affine functions of $\boldsymbol{\lambda}$, and the constraint in (17) is affine. So the dual problem (17) is always convex. And its optimal value, or the best lower bound of the primal problem (14), can be obtained in a rigorous manner by using, e.g., the projected subgradient method [40]. It can be shown [34] that a subgradient of function $g^\sigma$ at $\boldsymbol{\lambda}^\sigma$ is $\mathbf{x}_{\text{opt}}^\sigma$, i.e., the global optimizer of subproblem (18). In the projected subgradient method, at $t$ th iteration, the dual variables $\{\boldsymbol{\lambda}^\sigma\}$ are updated as

$$\boldsymbol{\lambda}^\sigma \leftarrow \mathcal{P}_\Lambda \left( \boldsymbol{\lambda}^\sigma + \alpha_t \mathbf{x}_{\text{opt}}^\sigma \right) \tag{19}$$

where $\mathcal{P}_\Lambda$ is the Euclidean projection onto set $\Lambda$. It is easy to show that the updating of $\lambda^\sigma$ in (19) reduces to the modification of $\theta^\sigma$ as follows

$$\theta_q^\sigma \mathrel{+}= \alpha_t \cdot \left( \mathbf{x}_{\text{opt}}^\sigma(q) - \frac{\sum_{\sigma \in \Upsilon(q)} \mathbf{x}_{\text{opt}}^\sigma(q)}{|\Upsilon(q)|} \right) \quad (20)$$

If the sequence $\{\alpha_t\}$ satisfies following conditions

$$\alpha_t \geq 0,\ \lim_{t \to \infty} \alpha_t = 0,\ \sum_{t=0}^{\infty} \alpha_t = \infty \quad (21)$$

the subgradient algorithm is guaranteed to converge to the optimal solution of problem (17) [34].

### 3.2 Energy minimization using DD

In this subsection, we will show how to design distributed algorithms to solve problems (2) and (3), using the DD technique described above. We propose two algorithms: L-DD (Linear DD) for minimizing energy function $E_1$, and Q-DD (Quadratic DD) for minimizing energy function $E_2$. We note here that various kinds of distributed algorithms for minimizing $E_1$ or $E_2$, or even more complex energy models, can be designed using the similar procedure as following based on DD technique. For example, in our previous work [43], the problem of minimizing $E_1$ was decomposed into $2N$ subproblems (two for each observation), and each subproblem was solved by direct searching.

*L-DD: Minimization of $E_1$*

Recall that the optimization variables $\mathbf{x} = \{x_q\}$ in problem (2) is used to specify the 0-1 state of the set of links $q \in A$. Let $Y_u$ be the set of observations made on camera $u$. We decompose the set $A$ into $2K$ overlapping subsets according to $Y_u$, where $K$ is the number of cameras in the networks. Specifically, $A$ is decomposed into $\Upsilon = \{A^-(Y_u), A^+(Y_u)\}_{u \in \{1:K\}}$, where $A^-(Y_u)$ (or $A^-(Y_u)$) is the set of links connecting observations in $Y_u$ to their possible predecessors (or successors). Note that each ordinary link is included exactly in two subset in $\Upsilon$, while each virtual link is included only in one subset. For each $u \in \{1, \ldots K\}$, we define two vectors of linking variables

$$\mathbf{x}^{u-} = \{x_q^{u-} \mid q \in A^-(Y_u)\} \text{ and } \mathbf{x}^{u+} = \{x_q^{u+} \mid q \in A^+(Y_u)\} \quad (22)$$

and parameter vectors

$$\boldsymbol{\theta}^{u-} = \left\{ \theta_q^{u-} \middle| \begin{array}{l} \theta_q^{u-} = \begin{cases} \tfrac{1}{2}\theta_q, & q \text{ is ordinary} \\ \theta_q, & q \text{ is virtual} \end{cases} \\ \forall q \in A^-(Y_u) \end{array} \right\} \text{ and } \boldsymbol{\theta}^{u+} = \left\{ \theta_q^{u+} \middle| \begin{array}{l} \theta_q^{u+} = \begin{cases} \tfrac{1}{2}\theta_q, & q \text{ is ordinary} \\ \theta_q, & q \text{ is virtual} \end{cases} \\ \forall q \in A^+(Y_u) \end{array} \right\} \quad (23)$$

where $\boldsymbol{\theta} = \{\theta_q\}_{q \in A}$ is the parameter vector in function $E_1$. The constraint sets associated with $\mathbf{x}^{u-}$ and $\mathbf{x}^{u+}$ are

$$\mathrm{X}^{u-} = \left\{ \mathbf{x}^{u-} \middle| \begin{array}{l} x_q^{u-} \in \{0,1\},\ \forall q \in A^-(Y_u) \\ \sum_{q \in A^-(y_i)} x_q^{u-} = 1,\ \forall y_i \in Y_u \end{array} \right\} \text{ and } \mathrm{X}^{u+} = \left\{ \mathbf{x}^{u+} \middle| \begin{array}{l} x_q^{u+} \in \{0,1\},\ \forall q \in A^+(Y_u) \\ \sum_{q \in A^+(y_i)} x_q^{u+} = 1,\ \forall y_i \in Y_u \end{array} \right\} \quad (24)$$

It is easy to verify that the set of $2K$ subproblems

$$\min_{\mathbf{x}^{u\pm} \in \mathrm{X}^{u\pm}} \langle \mathbf{x}^{u\pm}, \boldsymbol{\theta}^{u\pm} \rangle,\quad u \in \{1:K\} \quad (25)$$

satisfies conditions (12) and (13), and thus forms a decomposition of problem (2). Each subproblem can be viewed as a small sized linear assignment problem defined on a bipartite graph, see Fig.4. In the bipartite graph, nodes correspond to observations, and edges correspond to links. Each edge is associated with weight $\theta_q^{u-}$ or $\theta_q^{u+}$. The optimal assignment defines a 0-1 labeling of the links. We use Hungarian algorithm to find the global optimal labeling $\mathbf{x}_{\text{opt}}^{u\pm}$ for each graph.

The lower bound provided by (15) is maximized by adjusting the weight of edges in each bipartite graph iteratively using Eq.(20). Note that each ordinary link $q$ is included exactly in two bipartite graphs. According to Eq.(20), the weight of the corresponding edge is adjusted only when the labels of $q$ assigned by the two graphs are different, otherwise the weight remains untouched. The adjustments of edge weights encourage graphs to agree on their labeling of common links. As will be shown in Sec.3.3, complete agreement is guaranteed to be reached. Finally, the solution of problem (2) is obtained by traversing the labeling of links in all bipartite graphs. The L-DD algorithm is summarized in Alg.1.

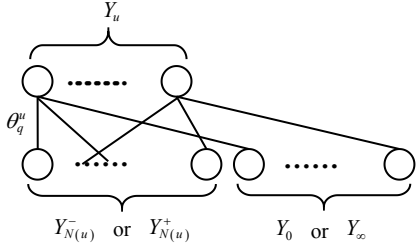

Fig.4 The bipartite graph for subproblem in L-DD. The nodes in the upper row represent the observations made on camera $u$, the nodes in the lower row represent possible predecessors (or successors) of $Y_u$, denoted as $Y_{n(u)}^-$ (or $Y_{n(u)}^+$), and virtual observations $Y_0$ or $Y_\infty$. The edges between the two disjoint set of nodes correspond to links, weighted by parameters $\theta_q$.

---

Alg.1: L-DD algorithm
    Input: observations $\{y_i\}$, parameters $\boldsymbol{\theta}^u$
    Output: optimal linking $\mathbf{x}_{opt}$
1:    Repeat
2:        camera $u = 1:K$ in parallel
3:           solve slave problems in Fig.4 using Hungarian algorithm
4:           adjust the weights of edges $\boldsymbol{\theta}^u$ by (20)
5:        end parallel
6:    Until convergence.

---

In L-DD algorithm, each camera $u$ is responsible for solving two slave problems $\{\mathbf{x}^{u\pm}, \boldsymbol{\theta}^{u\pm}, \mathbf{X}^{u\pm}\}$ located on it by using Hungarian algorithm. Parameter $\boldsymbol{\theta}^u$ is calculated based on observations generated on camera $u$, observations collected from its neighboring cameras, and the appearance and spatio-temporal similarity model stored on camera $u$. In each iteration, to adjust $\boldsymbol{\theta}^u$ an averaging operation is required which uses optimal solutions in the last iteration of two subproblems, one of which is located on $u$, the other is located on one of $u$'s neighbors. Thus, the L-DD algorithm can be implemented in a completely distributed manner.

*Q-DD: Minimization of $E_2$*

The optimization variables $\bar{\mathbf{x}} = \{x_q, x_{pq}\}$ of problem (3) are used to specify the 0-1 state of the set of links $q \in A$ and link pairs $pq \in B$. We decompose the set $A \cup B$ into $N$ overlapping subsets, $\Upsilon = \{A(y_i) \cup B(y_i)\}$, one for each observation. Here $N$ is the number of ordinary observations. Note that each ordinary link is included exactly in two subsets in $\Upsilon$, while each virtual link and each link pair is included in only one. For each $i \in \{1,\ldots,N\}$, we define a vector of linking variable

$$\bar{\mathbf{x}}^i = \{x_q^i, x_{pq}^i \mid q \in A(y_i), pq \in B(y_i)\} \tag{26}$$

and a corresponding vector of parameters $\bar{\boldsymbol{\theta}}^i$

$$\overline{\boldsymbol{\theta}}^i = \left\{ \theta_q^i, \theta_{pq}^i \middle| \begin{array}{l} \theta_q^i = \begin{cases} \frac{1}{2}\theta_q, & q \text{ is ordinary} \\ \theta_q, & q \text{ is virtual} \end{cases} \\ \theta_{pq}^i = \theta_{pq} \\ \forall q \in A(y_i), pq \in B(y_i) \end{array} \right\} \quad (27)$$

where $\overline{\boldsymbol{\theta}} = \{\theta_q, \theta_{pq}\}_{a \in A, pa \in B}$ is the parameters of energy function $E_2$. The constraint set that $\overline{\mathbf{x}}^i$ lies in is

$$\overline{X}^i = \left\{ \overline{\mathbf{x}}^i \middle| \begin{array}{l} x_q^i \in \{0,1\}, x_{pq}^i \in \{0,1\} \\ \sum_{p \in A^-(y_i)} x_p^i = \sum_{q \in A^+(y_i)} x_q^i = 1 \\ x_{pq} = x_p x_q \\ \forall q \in A(y_i), pq \in B(y_i) \end{array} \right\} \quad (28)$$

It is easy to verify that the set of $N$ subproblems

$$\min_{\overline{\mathbf{x}}^i \in \overline{X}^i} \langle \overline{\mathbf{x}}^i, \overline{\boldsymbol{\theta}}^i \rangle, \quad i \in \{1, \ldots, N\} \quad (29)$$

satisfies conditions (12) and (13), and thus forms a decomposition of problem (3). The subproblem (29) is illustrated in Fig.5. In the bipartite graph, the two disjoint sets of nodes represent incoming links to $y_i$ and outgoing links from $y_i$, respectively, and the edges between nodes correspond to pair of links. The parameters $\theta_q$ and $\theta_{pq}$ can be viewed as weights associated with nodes and edges, respectively. The optimal solution to subproblem $\overline{\mathbf{x}}^i_{opt}$ defines a 0-1 labeling of the nodes and edges in the bipartite graph. Note that the constraint (28) requires that only one of the edges, as well the associated pair of nodes, can be labeled as 1 (active). Thus, the subproblem (29) can be simply solved by direct searching in space $\overline{X}^i$, which contains only $|B(y_i)|$ elements.

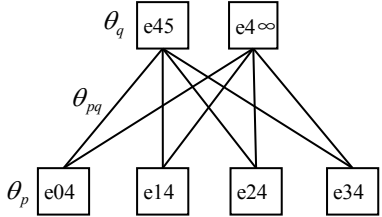

Fig.5 The bipartite graph for subproblem corresponding to $y_4$ in Fig.3. Here squar node represent links, e.g., e45 is the link between $y_4$ and $y_5$. The nodes e04 and e4∞ represent virtual links. The edges between nodes correspond to link pairs. Each node is associated with a parameter $\theta_a$, and each edge is associated with a parameter $\theta_{pa}$.

A lower bound of the optimal value of problem (3) is given by the sum of the optimal values of the $N$ subproblems in (29). The lower bound is maximized by adjusting the weights $\theta_q$ in each bipartite graph at each iteration. Note that each ordinary link $q$ is included in two bipartite graphs, while link pair $pq$ is included in only one graph. Thus the weight of $pq$ is never changed. If the labeling of common $q$ in two bipartite graphs conflict, the parameters $\theta_q$ are adjusted according to (20). The adjustments encourage subproblems to reach agreement on their labeling of common links. However, as we will show in Sec.3.3, it cannot be guaranteed that all conflicts are eliminated. In other words, there is a possible non-zero gap between the best lower bound provided by Q-DD algorithm and the optimal value of problem (3). Finally, based on the solutions to subproblems, the solution to problem (3) is determined as follows: upon convergence, if the subproblems agree on the labeling of common link $q$, then $x_q = x_q^i$; otherwise, we set $x_q = 0$ and $x_{q'} = 1$, where $q'$ is the virtual link belonging to $A^-(y_i)$ (if $x_q^i = 1$, $q \in A^-(y_i)$) or $A^+(y_i)$ (if $x_q^i = 1$, $q \in A^+(y_i)$). The variable $x_{pq}$ takes value according to $x_{pq} = x_p x_q$. The Q-DD algorithm is summarized as Alg.2.

In Q-DD algorithm, each camera $u$ is responsible to solve the subproblems corresponding to observations generated on it, by collecting observations generated on $u$'s neighbors. The averaging operation in (20) can also be performed locally, as it requires the solutions to subproblems located on neighboring cameras. Thus, Q-DD can be implemented in a completely distributed manner.

---

Q-DD algorithm: minimization of $E_2$
    Input: observations $\{y_i\}$ and parameters $\{\theta_q, \theta_{pq}\}$
    Output: optimal linking $\bar{\mathbf{x}}_{opt}$
1:     Repeat
2:         camera $u = 1:K$ in parallel
3:             solve slave problems in Fig.5
4:             adjust the weights of edges by (20)
5:         end parallel
6:     Until convergence.
7:     Determine the primal solution $\bar{\mathbf{x}}_{opt}$

---

### 3.3 Property of the lower bounds

Recall that in DD the original hard problem (primal problem) is treated by considering its dual, which is always convex and can be decomposed into several easy slave problems. The optimal value of the dual is a lower bound of the primal, and is maximized iteratively by subgradient method. If the duality gap is zero, the exact solution of the primal can be obtained based on optimizer of the dual; otherwise, only approximate solution can be found. In this subsection, we will discuss some properties of the lower bound provided by L-DD and Q-DD algorithms.

Lemma 1. The best lower bound computed by L-DD algorithm achieves the optimal value of problem (2).
Proof:
    In L-DD, problem (2) is decomposed as

$$\begin{cases} \min_{\{\mathbf{x}^u\}, \mathbf{x}} \sum_u \langle \mathbf{x}^u, \boldsymbol{\theta}^u \rangle \\ \text{s.t.} \quad \mathbf{x}^u \in X^u, \mathbf{x}^u = \mathbf{x}_{|u}, \forall u \in \{1:K\} \end{cases} \quad (30)$$

Here we use index $u$ instead of $u\pm$ to avoid notion clutters. We relax the consistence constraint $\mathbf{x}^u = \mathbf{x}_{|u}$ but keep the constraint $\mathbf{x}^u \in X^u$. By introducing a set of Lagrange multipliers $\boldsymbol{\lambda}^u$, the Lagrangian associated with problem (30) can be written as

$$L(\{\mathbf{x}^u\}, \mathbf{x}, \{\boldsymbol{\lambda}^u\}) = \sum_u \langle \mathbf{x}^u, \boldsymbol{\theta}^u \rangle + \sum_u \langle \boldsymbol{\lambda}^u, \mathbf{x}^u - \mathbf{x}_{|u} \rangle \quad (31)$$

The lower bound provided by L-DD is

$$\text{LB}_{\text{L-DD}} = \max_{\{\boldsymbol{\lambda}^u\}} \min_{\{\mathbf{x}^u \in X^u\}, \mathbf{x}} L(\{\mathbf{x}^u\}, \mathbf{x}, \{\boldsymbol{\lambda}^u\}) \quad (32)$$

$$= \max_{\{\boldsymbol{\lambda}^u\}} \min_{\{\mathbf{x}^u \in \mathcal{L}(X^u)\}, \mathbf{x}} L(\{\mathbf{x}^u\}, \mathbf{x}, \{\boldsymbol{\lambda}^u\}) \quad (33)$$

$$= \min_{\{\mathbf{x}^u \in \mathcal{L}(X^u)\}, \mathbf{x}} \max_{\{\boldsymbol{\lambda}^u\}} \sum_u \langle \mathbf{x}^u, \boldsymbol{\theta}^u \rangle + \sum_u \langle \boldsymbol{\lambda}^u, \mathbf{x}^u - \mathbf{x}_{|u} \rangle \quad (34)$$

$$= \min_{\{\mathbf{x}^u \in \mathcal{L}(X^u)\}, \{\mathbf{x}^u = \mathbf{x}_{|u}\}} \sum_u \langle \mathbf{x}^u, \boldsymbol{\theta}^u \rangle \quad (35)$$

$$= \min_{\mathbf{x} \in \mathcal{L}(X)} E_1(\mathbf{x}|\boldsymbol{\theta}) \quad (36)$$

$$= \min_{\mathbf{x} \in X} E_1(\mathbf{x}|\boldsymbol{\theta}) \quad (37)$$

Note that $\mathcal{L}(X^u)$ is a linear relaxation of $X^u$ which replaces the binary constraint $x_q \in \{0,1\}$ with linear constraint $x_q \geq 0$, and (33) results from the fact that the constraint matrix of $X^u$ is TUM and the Lagrangian $L$ is linear with respect to $x^u$. The exchange of min and max in (34) results from the strong duality. And $x$ is eliminated from (35) by dual feasibility. Equality (36) follows from decomposition conditions (12) and (13), where $\mathcal{L}(X)$ is a linear relaxation of $X$. Finally, we get (37) as $X$ is TUM and $E_1$ is linear. □

Lemma 1 states that L-DD is guaranteed to find the global optimizer of problem (2), just as many classical algorithms, e.g., Hungarian algorithm, MCF or KSP. However, these classical algorithms are designed for centralized computation. To the best of our knowledge, there is extremely little effort regarding the use of these classical algorithms to distributed applications. An interest exception is [41], in which the authors propose to use Belief Propagation (BP) algorithm to solve the networks flow problem. However, when apply it to the problem in this paper, a source node and a sink node, connecting to all other nodes corresponding to ordinary observations, are needed. Thus, communications between a central server on which the source and sink nodes are located, and cameras in the networks on which the ordinary nodes are located, are required. In contrast, L-DD only involves local computation on each camera and information exchange between neighboring cameras, making it very suitable in application of large scale smart camera networks.

Next, we will discuss the lower bound provided by Q-DD when apply it to problem (3). Unlike L-DD, Q-DD cannot be guaranteed to achieve the optimal value of problem (3), mainly due to the fact that the constraint in (3) is not TUM in general. We note a closely related algorithm based on Lagrangian relaxation was proposed recently in [23], which can also be used to minimizing quadratic energy function like $E_2$. To use [23], we rewrite problem (3) as

$$\begin{cases} \min_{\mathbf{z}} \sum_{pq \in B} c_{pq} z_{pq} \\ \text{s.t.} \quad z \in Z \end{cases} \quad (38)$$

where the constraint set is defined as

$$Z = \left\{ \{z_{pq}\} \middle| \begin{array}{l} \text{binary constraint: } z_{pq} \in \{0,1\}, \forall pq \in B \\ \text{flow conservation: } \sum_{p:pq \in B} z_{pq} = \sum_{r:qr \in B} z_{qr}, \forall q \in \overline{A} \\ \text{uniqueness constraint: } \sum_{pq \in B(y_i)} z_{pq} = 1, \forall i \in \{1:N\} \end{array} \right\} \quad (39)$$

In (38), the optimization variables are considered as flows between links $p$ and $q$. The set $\overline{A}$ is the subset of $A$ which contains all ordinary links in $A$. The flow conservation constraint amounts to requiring the consistence between subproblems defined as in Q-DD. It is easy to verify that problem (38) is equivalent to (3) if the coefficients in the cost function are defined as $c_{pq} = \frac{1}{2}\theta_p + \frac{1}{2}\theta_q + \theta_{pq}$ (if $q$ is virtual, $\frac{1}{2}\theta_q$ should be replaced by $\theta_q$). In [23], the uniqueness constraint is relaxed and the Lagrangian is defined as

$$L(\mathbf{z}, \boldsymbol{\lambda}) = \sum_{pq \in B} c_{pq} z_{pq} + \sum_{i=1}^{N} \lambda_i \left( \sum_{pq \in B(y_i)} z_{pq} - 1 \right) \quad (40)$$

For a fixed $\boldsymbol{\lambda}$, minimizing the Lagrangian $L$ with respect to $\mathbf{z}$ under the binary and flow conservation constraint

$$\min_{\mathbf{z} \in M} L(\mathbf{z}, \boldsymbol{\lambda}) \quad (41)$$

is a standard min-cost flow problem, which can be solved efficiently by classical MCF or KSP algorithms. Here we use $M$ to denote the set of $\mathbf{z}$ satisfying the binary and flow conservation constraints. The resulting lower bound on the cost in (38) is maximized iteratively by updating $\boldsymbol{\lambda}$ using a subgradient method, such as

$$\lambda_i^{(t)} {+}= \alpha_t \left( \sum_{pq \in B(y_i)} \mathbf{z}_{\text{opt}}^{(t-1)}(pq) - 1 \right), i \in \{1:N\} \tag{42}$$

where $\mathbf{z}_{\text{opt}}^{(t-1)}$ is the optimizer of (41) obtained in the last iteration.

Lemma 2. The best lower bound computed by Q-DD is the same as that of [23].

Proof:

The lower bound provided by [23] is

$$\text{LB}_{\text{Butt}} = \max_{\boldsymbol{\lambda}} \min_{\mathbf{z} \in M} L(\mathbf{z}, \boldsymbol{\lambda}) \tag{43}$$

$$= \max_{\boldsymbol{\lambda}} \min_{\mathbf{z} \in \mathcal{L}(M)} L(\mathbf{z}, \boldsymbol{\lambda}) \tag{44}$$

$$= \min_{\mathbf{z} \in \mathcal{L}(M)} \max_{\boldsymbol{\lambda}} \sum_{pq \in B} c_{pq} z_{pq} + \sum_{i=1}^{N} \lambda_i \left( \sum_{pq \in B(y_i)} z_{pq} - 1 \right) \tag{45}$$

$$= \min_{\mathbf{z} \in \mathcal{L}(Z)} \sum_{pq \in B} c_{pq} z_{pq} \tag{46}$$

$$\leq \min_{\mathbf{z} \in Z} \sum_{pq \in B} c_{pq} z_{pq} \tag{47}$$

We use $\mathcal{L}(M)$ and $\mathcal{L}(Z)$ to denote the linear relaxation of $M$ and $Z$. Equality (44) is true because constraint $M$ is TUM and $L$ is linear with respect to $\mathbf{x}$. We get the inequality (47) because

$$\mathcal{L}(Z) \supseteq \text{convexhull}(Z) \tag{48}$$

Now let's look at Q-DD. In Q-DD, problem (3) is transformed into

$$\begin{cases} \min_{\{\overline{\mathbf{x}}^i\}, \overline{\mathbf{x}}} \sum_{i=1}^{N} \langle \overline{\mathbf{x}}^i, \overline{\boldsymbol{\theta}}^i \rangle \\ \text{s.t.} \quad \overline{\mathbf{x}}^i \in \overline{X}^i, \overline{\mathbf{x}}^i = \overline{\mathbf{x}}_{|i}, \forall i \in \{1:N\} \end{cases} \tag{49}$$

We relax the consistence constraint, but keep the constraint $\overline{\mathbf{x}}^i \in \overline{X}^i$ in (49), and write the Lagrangian as

$$L(\{\overline{\mathbf{x}}^i\}, \{\lambda_q\}) = \sum_{i=1}^{N} \left( \sum_{q \in A(y_i)} \theta_q^i x_q^i + \sum_{pq \in B(y_i)} \theta_{pq}^i x_{pq}^i \right) + \sum_{q \in A} \lambda_q \left( x_q^{i:q \in A^-(y_i)} - x_q^{j:q \in A^+(y_j)} \right) \tag{50}$$

The lower bound provided by Q-DD is

$$\text{LB}_{\text{Q-DD}} = \max_{\{\lambda_q\}} \min_{\{\overline{\mathbf{x}}^i \in \overline{X}^i\}} L(\{\overline{\mathbf{x}}^i\}, \{\lambda_q\}) \tag{51}$$

$$= \max_{\{\lambda_q\}} \min_{\mathbf{z} \in H} \sum_{i=1}^{N} \sum_{pq \in B(y_i)} c_{pq} z_{pq} + \sum_{q \in A} \lambda_q \left( \sum_{p:pq \in B(y_i)} z_{pq} - \sum_{r:qr \in B(y_j)} z_{qr} \right) \tag{52}$$

$$= \max_{\{\lambda_q\}} \min_{\mathcal{L}(H)} \sum_{i=1}^{N} \sum_{pq \in B(y_i)} c_{pq} z_{pq} + \sum_{q \in A} \lambda_q \left( \sum_{p:pq \in B(y_i)} z_{pq} - \sum_{r:qr \in B(y_j)} z_{qr} \right) \tag{53}$$

$$= \min_{\mathcal{L}(H)} \max_{\{\lambda_q\}} \sum_{i=1}^{N} \sum_{pq \in B(y_i)} c_{pq} z_{pq} + \sum_{q \in A} \lambda_q \left( \sum_{p:pq \in B(y_i)} z_{pq} - \sum_{r:qr \in B(y_j)} z_{qr} \right) \tag{54}$$

$$= \min_{\mathcal{L}(Z)} \sum_{i=1}^{N} \sum_{pq \in B(y_i)} c_{pq} z_{pq} \tag{55}$$

$$\leq \min_{\mathbf{z} \in Z} \sum_{pq \in B} c_{pq} z_{pq} \tag{56}$$

Here we use $H$ to denote the set of $\mathbf{z}$ satisfying the binary and uniqueness constraints. Equality (52) follows from the fact that there is a one-to-one correspondence between $\{\overline{\mathbf{x}}^i \in \overline{X}^i\}$ and $\mathbf{z} \in H$, and for each corresponding pair, the following equalities

$$\sum_{q \in A(y_i)} \theta_q^i x_q^i + \sum_{pq \in B(y_i)} \theta_{pq}^i x_{pq}^i = \sum_{pq \in B(y_i)} c_{pq} z_{pq} \tag{57}$$

and

$$\sum_{p:pq\in B(y_i)} z_{pq} = x_q^i, \quad \sum_{r:qr\in B(y_j)} z_{qr} = x_q^j \tag{58}$$

hold true. We get (53) because that $H$ is TUM. It follows trivially from (46) and (55) that $LB_{Q\text{-}DD} = LB_{Butt}$. □

Lemma 2 states that both Q-DD algorithm and [23] can find the optimal value of the same linear relaxation of the original problem (3). However, [23] is a centralized method which originally used for multi-target tracking in videos captured by a single camera. In contrast, Q-DD can be implemented in a distributed manner, making it very suitable in camera network application. In addition, it seems difficult to extend [23] to higher (than quadratic) order model which can evaluate the quality of tracklets consisting more than three observations. But distributed algorithms for optimizing higher order energy model can be easily designed based on DD technique.

4. Experiments

4.1 Settings

We evaluate the proposed method in two experiments of pedestrian tracking using disjoint cameras networks. One is performed in a campus garden, and the other is in an office building, as shown in Fig.6 (a) and (c). In each experiment, there are ten cameras mounted on different sites for collecting video data. The constraints on the movement of pedestrians imposed by the geography of monitored regions are encoded by the camera networks topology, as shown in Fig.6 (b) and (d). To focus on data association algorithms evaluation, we extract "observations" as described in Sec.2 from video data manually. Specifically, we segment pedestrians from video frames by hand and divide a person image into lower and upper regions. For each region, the 256 dimensional histogram vectors in RGB channels are computed and used as original appearance features. We also record the time instants and directions of person entering into or leaving from a camera's FOV, and use them as spatio-temporal features.

*Scenario 1: Campus Garden experiment.* The camera networks used in this experiment consists of ten cameras mounted in our campus garden, and the layout and corresponding topology are shown in Fig.6 (a) and (b). During the experiment, 10 persons present in the monitored region. We gather altogether 300 observations from 90-minites video collected by the camera networks. In outdoor scenario, the lighting conditions at each site are similar, but the monitored region is larger and the distance between cameras is longer than that of indoor case.

*Scenario 2: Office Building experiment.* The experiment in this scenario was conducted with ten cameras mounted in an office building, five in the first floor and five in the second floor. The camera layout and the corresponding topology are shown in Fig.6 (b) and (d). Altogether 300 observations originated from 10 persons are extracted from the 70-minites video data collected by the cameras in the networks. The main challenges are the significant variations in lighting conditions and view angles. For example, the areas covered by camera B and E are clearly dim due to the lack of lighting. And the view angles at stairs, C and F, are quite different from those of other sites.

4.2 Criteria

We use the following measures to evaluate the algorithms: the estimated number of objects $K$, the precision $P$, recall $R$, and F-measure of the reconstructed trajectories. Let $\{Y_1^*,...,Y_{K^*}^*\}$ be the ground truth, each of which is a set of observations that belong to the same object. Let $\{Y_1,...,Y_K\}$ denote the mutually disjoint subsets of $Y$ given by data association algorithms. Each subset represents the reconstructed trajectory of a single person. The precision and recall are

defined as

$$P = \frac{1}{K}\sum_{i=1}^{K}\max_{j}\frac{|Y_i \cap Y_j^*|}{|Y_i|}, \qquad R = \frac{1}{K^*}\sum_{j=1}^{K^*}\max_{i}\frac{|Y_i \cap Y_j^*|}{|Y_j^*|}, \qquad F = \frac{2\cdot P\cdot R}{P+R} \qquad (59)$$

The precision and recall represent the fidelity and the completeness of the estimated trajectories, respectively. A reconstructed trajectory with a single observation has a 100% precision, and a reconstructed trajectory with all observations has a 100% recall. And the F-measure is the harmonic mean of these two measures.

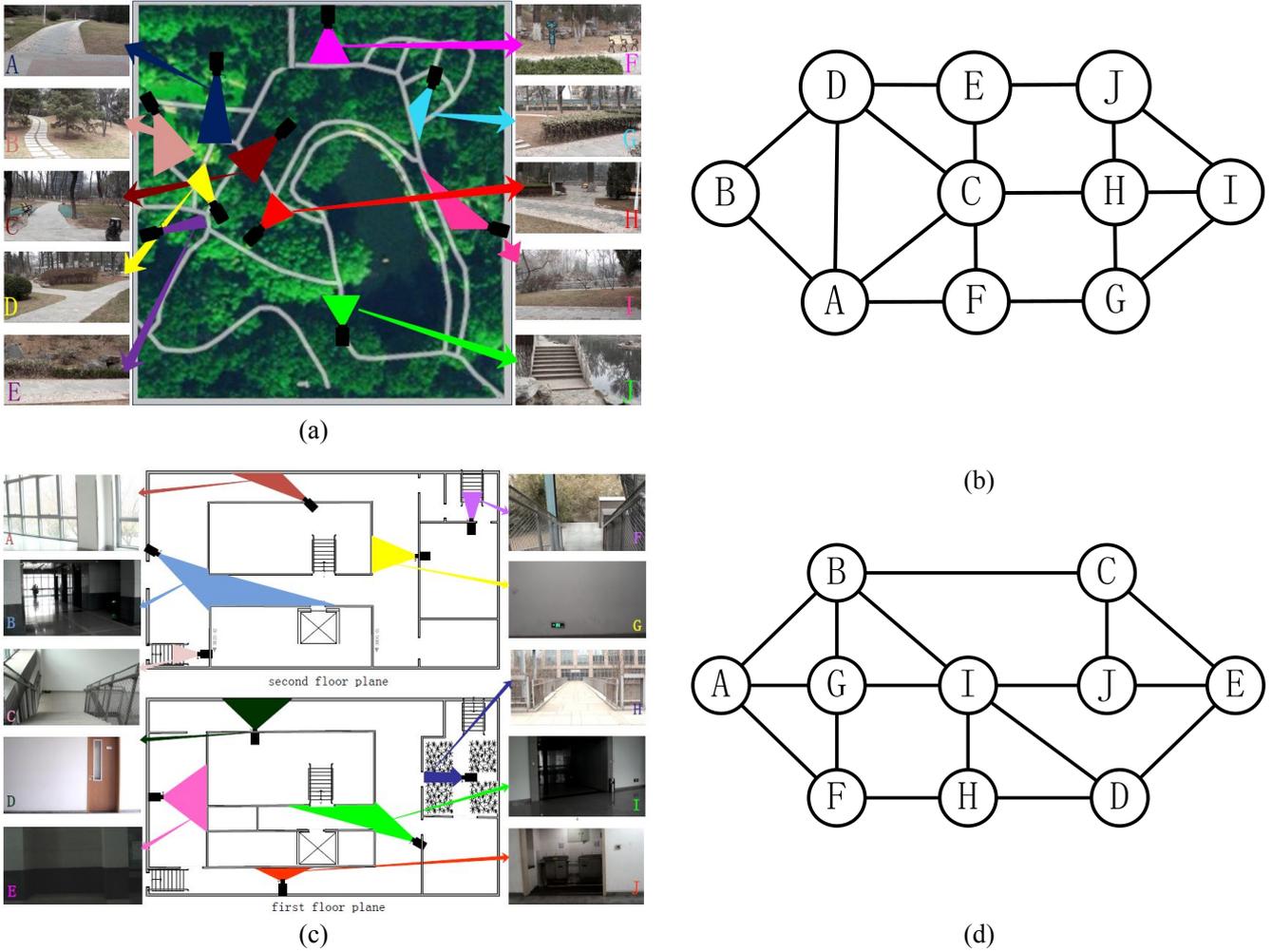

**Fig.4 Experiment settings. (a) Campus garden layout. (c) Office building layout. (b) and (d) Corresponding topology.**

To evaluate the speed of the algorithms, we measure the execution time $\tau(Y)$ of each algorithm implemented in Matlab on common PC. The time cost involved with data collection, communication, object segmentation and tracking of object on single camera are ignored in this paper. Note that for distributed algorithms, the actual running time in real applications can be much shorter than $\tau(Y)$ because several agents can perform computation in parallel.

4.3 Results

We apply our proposed algorithms and some of the state of the art methods to the above two datasets. Table 1 lists the algorithms used for comparison. These algorithms differ from one another in the variable and model they used, the property of solution, and the implementation manner. Table 2 and 3 show the experimental results, including the association accuracy and running time. Fig.8 and Fig.9 show selected trajectories recovered by our methods in the experiments. Before conducting experiments, the observations extracted from about five hour-long video data collected by the camera networks are manually labeled and used for learning the CBTF mapping. Then we transfer the appearance measurement by CBTF firstly and apply Eq.(6) to calculate the appearance similarity. The spatio-temporal similarity is calculated using Eq.(7), and the transition probability between neighboring cameras are set according to prior knowledge. Note here we use a weak but more realistic uniform model (8) instead of Gaussian model in previous works [1,12,32,43]. We set $\Delta_{i,j}^{\min} = 0.25 \times \Delta_{i,j}^{mean}$ and $\Delta_{i,j}^{\max} = 4 \times \Delta_{i,j}^{mean}$, where $\Delta_{i,j}^{mean}$ is the mean traveling time between camera i and j, which are learnt from training data.

Table 1 Algorithms for comparison

|   | model | variable | solution | implementation |
|---|---|---|---|---|
| C-inf [1] | linear | lable of observation | marginal probability of label, approximate | centralized, online |
| D-inf [32] | linear | lable of observation | marginal probability of label, approximate | distributed, online |
| MHT [17] | linear | lable of observation | MAP labeling, approximate | centralized, online |
| MCF [21] | linear | flow of link | optimal linking, exact | centralized, offline |
| D-opt [33] | linear | partition of observations | optimal linking, approximate | distributed, offline |
| L-DD | linear | 0-1 state of link | optimal linking, exact | distributed, offline |
| LR-MCF [23] | quadratic | 0-1 state of link pair | optimal linking, approximate | centralized, offline |
| Q-DD | quadratic | 0-1 state of link & link pair | optimal linking, approximate | distributed, offline |

Table 2 Results on Campus garden dataset

|   | precision (%) | recall (%) | F-measure (%) | num | time (s) |
|---|---|---|---|---|---|
| C-inf | 74.18 | 83.33 | 78.50 | 15 | 291300 |
| D-inf | 74.70 | 75.48 | 75.09 | 13 | 24 |
| MHT | 88.65 | 85.24 | 86.91 | 12 | 8650 |
| MCF | 93.00 | 92.67 | 92.83 | 10 | 26 |
| D-opt | 90.34 | 86.00 | 88.12 | 14 | 61 |
| L-DD | 93.00 | 92.67 | 92.83 | 10 | 416 |
| LR-MCF | 100 | 90.33 | 94.92 | 12 | 19874 |
| Q-DD | 100 | 90.33 | 94.92 | 12 | 688 |

Table 3 Results on Office building dataset

|   | precision (%) | recall (%) | F-measure (%) | num | time (s) |
|---|---|---|---|---|---|
| C-inf | 47.27 | 46.33 | 46.80 | 12 | 231580 |
| D-inf | 66.81 | 61.00 | 63.77 | 10 | 30 |
| MHT | 69.81 | 65.00 | 67.32 | 11 | 9832 |
| MCF | 87.70 | 86.33 | 87.01 | 10 | 37 |
| D-opt | 82.23 | 77.67 | 79.88 | 15 | 54 |
| L-DD | 87.70 | 86.33 | 87.01 | 10 | 340 |
| LR-MCF | 98.57 | 98.33 | 98.45 | 10 | 33296 |
| Q-DD | 98.57 | 98.33 | 98.45 | 10 | 1034 |

*A. Our methods*

We implement L-DD algorithm to solve the problem (2). The cost associated with linking variable is calculated as described in Sec.2.3. We set the cost of virtual links to 25 in both experiments. We find that our algorithm is insensible to this value within the range of 20~50. In Campus garden experiment, there are altogether 17313 linking variables. In Office building experiment, the set of linking variable contains 18910 elements. In both experiment, the problem (2) is decomposed into 20 linear assignment subproblems (two for each camera), which are solved by Hungarian algorithm. The sequence $\{\alpha_t\}$ in Eq.(18) is chosen as $5/\sqrt{t}$. Fig.6 shows the energy curves of L-DD in Campus garden and Office building experiments, respectively. At each iteration, the dual energy is the sum of optimal values of all subproblems. To calculate the primal energy, we need to determine the solution to primal problem (i.e. problem (2)) from the solutions to subproblems. We find the primal solution by eliminating the conflicts between solutions of subproblems. Note that each ordinary link is involved in exactly two subproblems. If the solutions to the two subproblems disagree on the state of some ordinary link $q$, we set $x_q = 0$ and set the proper virtual link to be 1 to make the uniqueness constraint to be satisfied. Once the primal solution is determined, the primal energy is calculated as $E_1(\mathbf{x}|\boldsymbol{\theta})$. It can be seen from Fig.6 that the primal and dual energy reach the same value of 1166 after 1140 iterations in Campus garden experiment and the value of 2208 after 1003 iterations in Office building experiment, implying that the global optimal solution of problem (2) are found. The L-DD algorithm gives the highest F-measure among all linear model based methods for comparison in the two experiments. The running time of L-DD includes the time for calculating the link cost, say, 24 and 35 seconds, for the two experiments respectively.

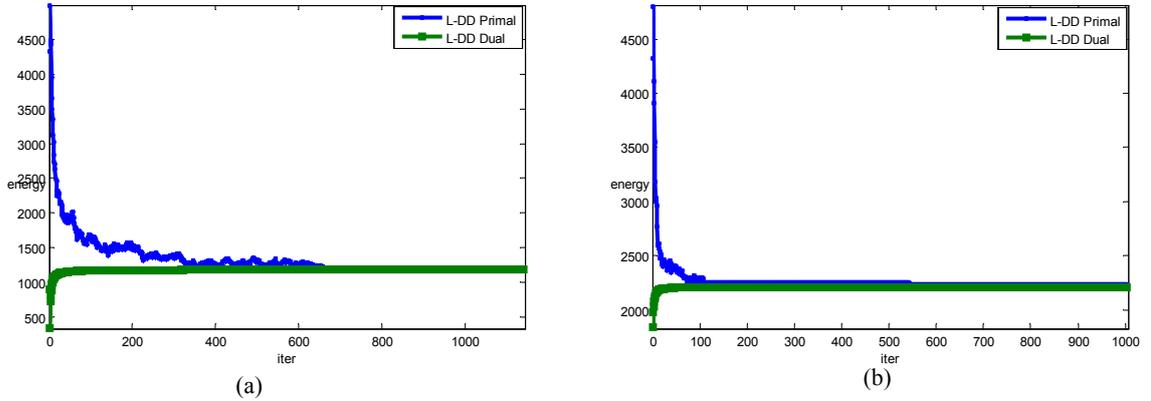

**Fig.6 Energy curves of L-DD. (a) Campus Garden; (b) Office Building.**

We also implement Q-DD algorithm to solve problem (3). The costs associated with links and link pairs are calculated as described in Sec.2.3. For calculating $\theta_{pq}$, we only use the appearance constraint, enforcing appearance similarity between predecessor and successor of a single observation, but ignore the higher-order spatio-temporal constraint. There are extremely large number of optimization variables in problem (3), say, 652768 for Campus garden and 777940 for Office building, respectively. In Q-DD, the primal problem (3) is decomposed into 300 subproblems (one for each observation) for both experiments, which are solved by direct searching. We note that in our experiments, a single subproblem contains at most 5180 optimization variables and solving it requires only 5180 searching operations (because of the uniqueness constraint). In Q-DD, the coefficients sequence $\{\alpha_t\}$ for updating the cost is chosen as $5/\sqrt{t}$. The energy curves are shown in Fig.7. The primal solution is obtained from solutions to subproblems by setting the label of conflicting link to be 0, and the proper virtual link to be 1 to fulfill the uniqueness constraint. We can see from Fig.7 that in our experiments, upon convergence the primal and dual energies reach the same value (although this cannot be guaranteed theoretically), say, 4081 in Campus garden after 506 iterations and 6241 in Office building after 1063 iterations. Because of the effective use of higher order constraint, the Q-DD achieves the best result in both experiments among all the methods for comparison, as shown in Table 1. The running time of Q-DD includes the time for calculating the costs of links and link pairs, say, 536 and 736 second, for the two experiments respectively.

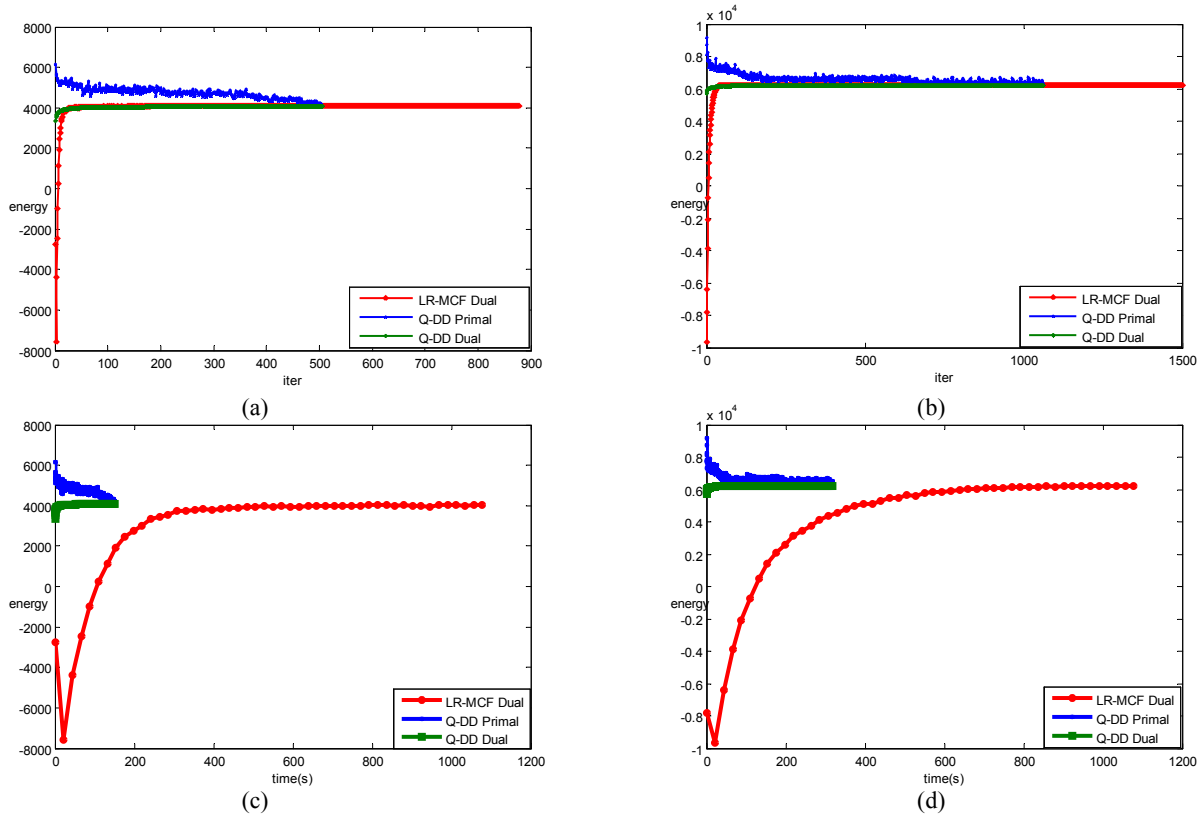

**Fig.7 Energy curves of Q-DD and LRMCF. (a) Campus Garden; (b) Office Building; (c) and (d) are obtained by calibrating the x-axis of (a) and (b) into real time. .**

*B. Comparison with C-inf*

We first compare our methods with C-inf [1]. In C-inf, a labeling variable is assigned to each observation and the posterior distribution of each labeling variable is calculated using the Bayesian rule. The posterior is conditioned on all information made in the camera networks up to the current time step, so the inference can be performed online. In C-inf, the joint distribution of observations and hidden variables (including labeling variable, counting variable and appearance model parameters) is encoded by a hybrid dynamic Bayesian networks, and the marginal of labeling and counting variables are inferred by using the approximate Assumed Density Filter (ADF) [19]. By inferring the counting variables, the number of persons can be estimated from observations automatically. In C-inf, the appearance feature is taken as a 6-D vector containing color means (RGB) computed over the lower and upper region of a person's image, and the appearance of a person under different camera sites is assumed to follow a single Gaussian model with Normal-Inverse Wishart distributed mean and covariance parameters, allowing the appearance model to be updated elegantly during inference. The main drawback of C-inf is its speed. We find that the computational and memory requirement of C-inf increases rapidly with the number of observations. To make it tractable on our datasets, we limit the memory depth of the central server to be 25 and set the maximum value of counting variable to be 20 in both experiments. It is obvious in Table 2 and 3 that the speed of C-inf is much slower than ours, mainly due to the combinatory explosion of sampling space of hidden variables and the centralized nature of the algorithm. In addition, we also note that C-inf does not show superiority to other algorithm in terms of association accuracy, especially in the case of Office building. This can be attributed to the unrealistic appearance model and the truncation of the older observations during inference. In fact, for centralized inference, a much deeper memory stack is required to ensure that the true predecessor of the current observation is reserved. But this will lead to unacceptable computational and memory cost.

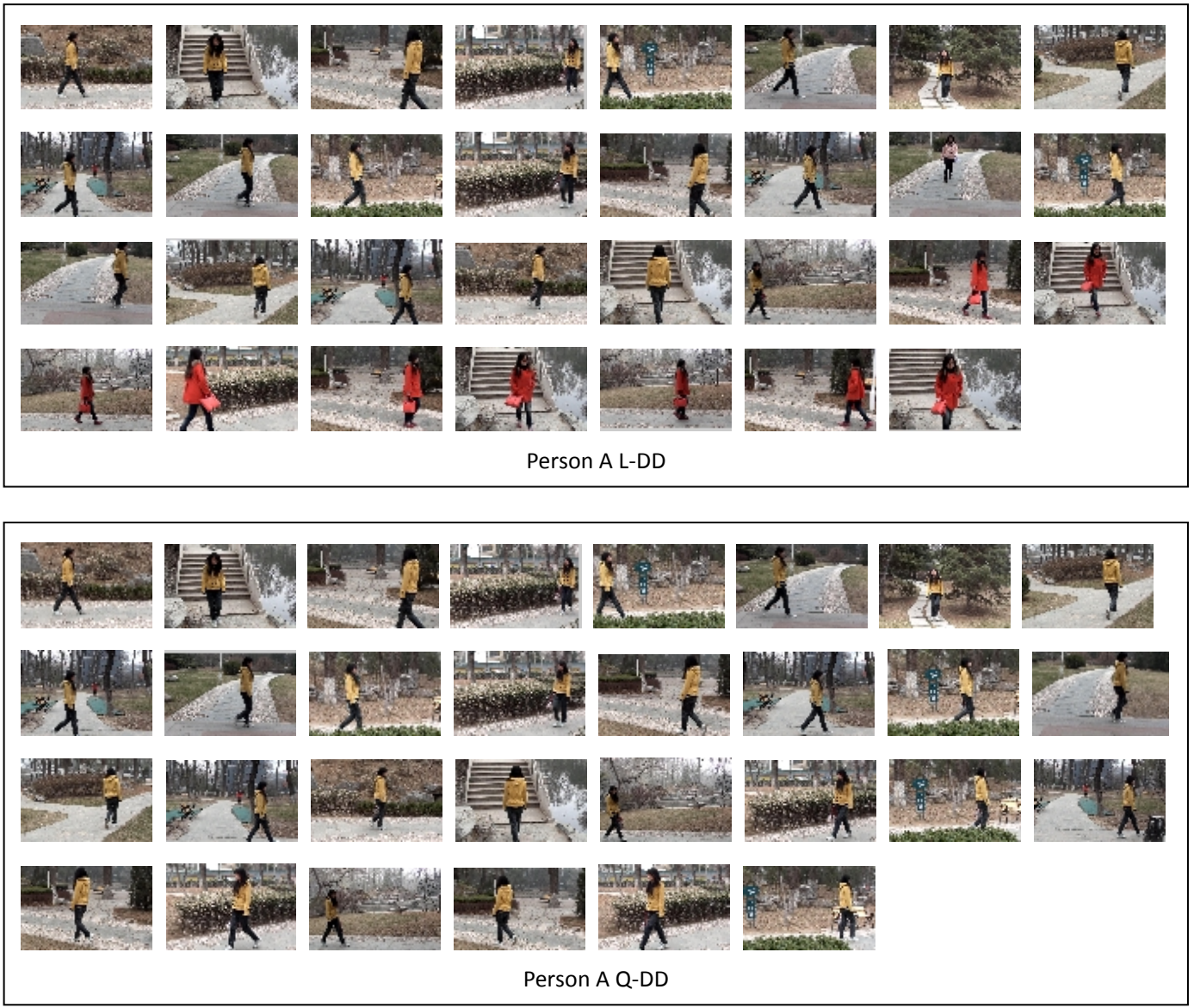

Fig.8 Recovered trajectories in Campus garden experiment.

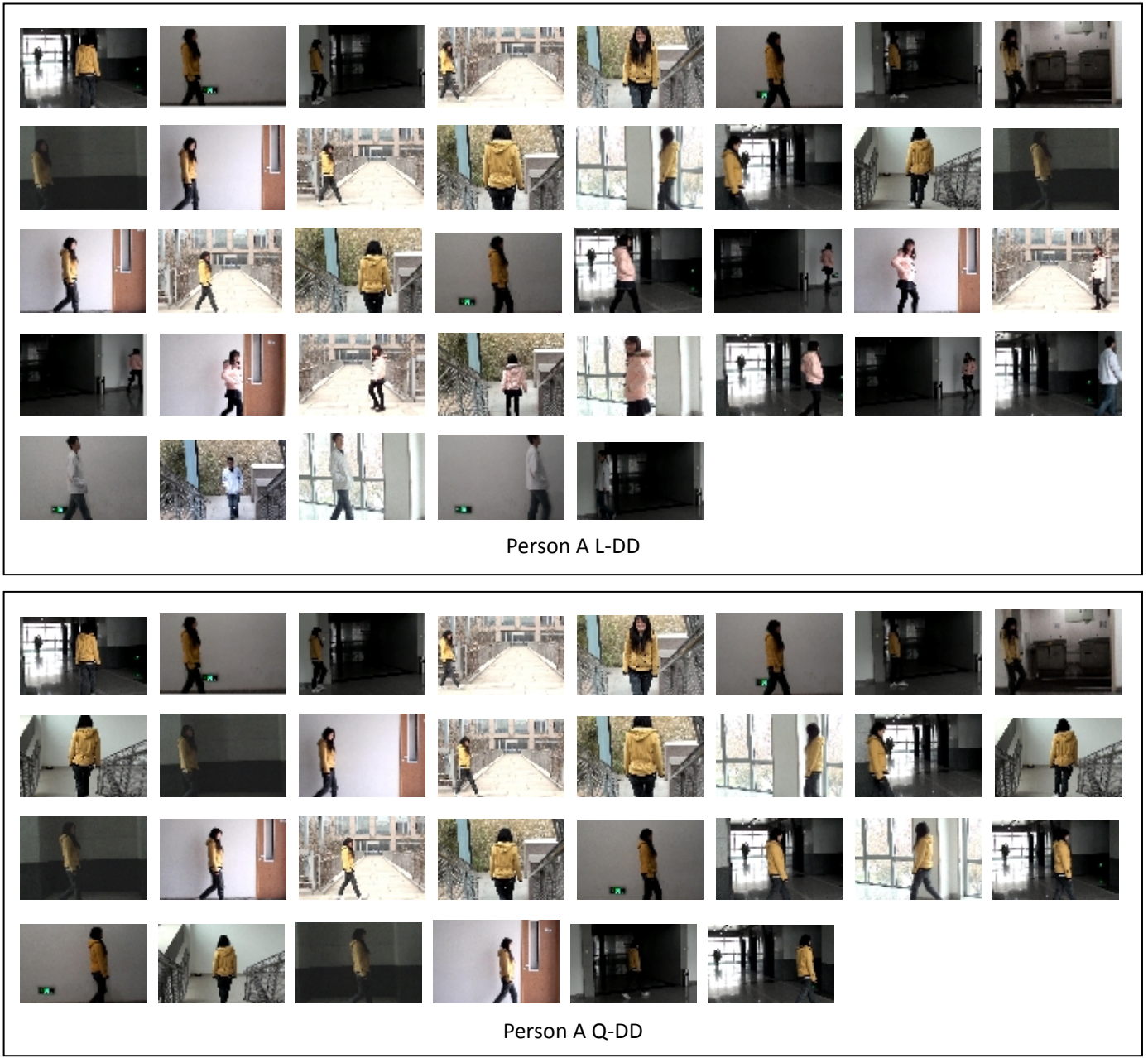

Fig.9 Recovered trajectories in Office building experiment.

## C. Comparison with. D-inf

We also compare our methods with D-inf [32], which improves on C-inf in the following two aspects. Firstly, in D-inf, the inference is based solely on local computation on each camera and information exchange between neighboring cameras, which results in a dramatic reduction of computational burden. Secondly, in D-inf, the Gaussian assumption on appearance used in C-inf is discarded. Instead, the similarity measure between observation and its predecessors are computed and used for defining the likelihood. We refer the interested readers to [32] for details. In our implementation, the appearance and spatio-temporal feature are chosen as described in Sec.4.1, and the similarity measure is calculated using the same linear model as in L-DD as describe in Sec.2.3. The memory depth of each camera is set to be 25 and the maximum value of counting variable is set to be 20 in both experiments. From Table 2 and 3 we can see that D-inf runs much faster than C-inf, and outperforms C-inf in terms of accuracy in the Office building experiment, where the observing condition varies significantly across different cameras. However, the accuracy of D-inf is still lower than that of our methods. This can be mainly attributed to the approximation made in D-inf. Specifically, in D-inf, to make the computation tractable, during inference the labeling variables are assumed to be marginally independent and the joint distribution is approximated by the product of marginal of individual labeling variables. Besides, in D-inf the inference is performed online and the posterior of labeling variable is calculated based only on history information. In contrast, by exploiting the special structure of the problem, L-DD (and MCF) can find the exact global optimal solution efficiently with respect to the same linear model underlying D-inf. Although Q-DD can not be guaranteed to find the global optimal solution because the quadratic model is used, it still gives the highest accuracy because of the power of higher order constraint.

*D. Comparison with MHT*

MHT is a classical technique for multi-targets tracking. Recently it has been introduced to tracking persons in camera networks with non-overlapping FOVs [17]. We implement MHT by constructing and maintaining a hypothesis tree. We use $\Omega_k^i$ to denote $k$ th hypothesis at time step $i$ when observation $y_i$ is generating. Each hypothesis $\Omega_k^i$ is a specific labeling of observations $y_{1:i}$, and the labeling variable $x_i$ indicates which object the observation $y_i$ comes from. Each hypothesis has a probability of being correct, which is evaluated as the joint probability $p(x_{1:i}|y_{1:i})$. At time step $i+1$, the hypotheses set $\Omega^i$ is used to produce hypotheses $\Omega^{i+1}$. For each hypothesis $\Omega_k^i$, a new set of children hypotheses $^k\Omega^{i+1}$ is generated by enumerating the possible values of $x_{i+1}$, and the probability of hypothesis $^k\Omega_l^{i+1}$ being correct is calculated as

$$p(x_{1:i+1}|y_{1:i+1}) = p(x_{i+1}|x_{1:i}, y_{1:i+1}) p(x_{1:i}|y_{1:i}) \tag{60}$$

The conditional probability $p(x_{i+1}|x_{1:i}, y_{1:i+1})$ can be computed by using the similarity measure between $y_{i+1}$ and previous observations. For example, suppose that $x_{i+1} = x_m$ and $x_{i+1} \neq x_n, \forall n \in [m+1, i]$, then

$$p(x_{i+1}|x_{1:i}, y_{1:i+1}) \propto \phi(y_{i+1}, y_m) \prod_{n=m+1}^{i} (1 - \phi(y_{i+1}, y_n)) \tag{61}$$

where the similarity measure $\phi$ are calculated as in Sec.2.3. Note that $m = 0$ implies $y_{i+1}$ is originated from a new object. And we define $\phi(y_{i+1}, y_0)$ as the similarity measure corresponding to the virtual link. The hypothesis tree constructed in this way will grow exponentially with time. We control the number of hypotheses by pruning the tree at each time step. Specifically, we set a maximum number $M$ of leaves of the tree at each time step, and delete the extra leaves with lower probability of being correct. The results in Table 2 and 3 are obtained by setting $M = 30$. We find that the accuracy of MHT is inferior to L-DD and MCF, mainly due to its online and approximation nature. Furthermore, as we can see from the table, MHT is much slower than our methods. Choosing a smaller $M$ can improve its speed, but at the cost of accuracy. For example, when $M = 10$, the running time of MHT in Campus Garden experiment is 3620s, but the F-measure drops to 74.29%.

*E. Comparison with MCF*

MCF was originally proposed for multi-targets tracking in a single camera view [21], and recently has been used for data association across disjoint camera views [16]. In our implementation, the cost-flow network is defined as that in Fig.3(b), augmented with a source and a sink node, both connecting to all observation nodes in the network. A cost $\theta$ as defined in Sec.2.3 is assigned to each arc in the network, and the lower and upper bound of flow on each arc are set to be 0 and 1, respectively. Given the amount of flow, i.e. the number of person's trajectories, sent from the source node to the sink node, the algorithm in [22] can be used to solve for the flow paths with the minimal total cost. However, there is some subtleness in the implementation of MCF, and we would like to record it here. We note that all cost $\theta$ are positive, so directly applying the algorithm in [22] to our problem would give a trivial zero solution. Thus a negative bias should be added to the cost of each arc. Initially, we choose the bias as $B \geq \max\{\theta_q\}$ and let the cost of each arc to be $\theta_q - B$, ensuring all cost to be negative. But we find that in the solution under this setting, a flow path corresponding to a single person's trajectory tends to split into two or more sub-paths. This phenomenon puzzled us for a lot of time, but finally we figure out the reason. Suppose that a flow path is replaced by two flow paths by breaking it at an ordinary arc $p$, the total cost change would be

$$-(\theta_p - B) + 2(\theta_0 - B) = 2\theta_0 - B - \theta_p \qquad (62)$$

Here ordinary arc means arc connecting two observation nodes, and virtual arc means arc connecting observation node to source or sink node, and $\theta_0$ is the cost of virtual arc. Unless $\theta_p$ is small enough, or the linked observations are similar enough, to make sure that $\theta_p < 2\theta_0 - B$, the total cost change would be negative. Consequently MCF would break the flow path at arc $p$ into two pieces. As the bias is chosen as $B \geq \max\{\theta_q\}$, many links between two observations of a single persons would break. To cope with this problem, finally we set the cost of ordinary links as $\theta_q - B$, but the cost of virtual links as $\theta_0 - \frac{B}{2}$. From Table 2 and 3 we can see that MCF (and also L-DD) achieves the best results among all linear model based methods in both experiments, and runs very fast. If the costs of links are calculated beforehand, it only need 2~3 seconds for MCF to find the optimal linking. But we note here that we use the highly optimized C code provided by the author [23]. Furthermore, MCF is a centralized method in which all observations need to be collected to a central server for processing. This makes MCF unsuitable for large scale applications. In fact, the complexity of the cost-flow network grows rapidly with the number of observations. In addition, to determine the number of trajectory, in MCF the optimal cost should be calculated over all possible amount of flow sent from source to sink. In other words, MCF needs to search in the space of number of persons. If the number of persons under tracking was increased, the running time of MCF would be longer. In contrast, the running time of L-DD or Q-DD is independent from the number of persons under tracking.

## F. Comparison with D-opt

In [33] a distributed optimization based data association algorithm (D-opt) is proposed, which is applicable to large scale camera networks. Similar to our method, D-opt decomposes the original problem into several subproblems, one for each camera, of grouping observations made in each camera's neighborhood into local trajectories. Each subproblem is formulated as an observation-predecessor assignment problem and solved by using the Hungarian algorithm. A voting algorithm is employed to resolve the possible conflict between solutions of different subproblems, ensuring that each observation has a unique predecessor. From Table 2 and 3 we can see that D-opt runs very fast. However, in experiments we find that D-opt may link a single observation to more than one successors. A typical case is shown in Fig.10, by solving the subproblem on camera A, observation O2 is assigned to its assumed predecessor O1. Independently, camera C thinks that the predecessor of O3 should be O1. There is no conflict between camera A and C because each observation has a unique predecessor. But observation O1 has two successors and the trajectory of that person bifurcates at O1. This phenomenon results in more estimated trajectories than the ground truth, as shown in Table 2 and Table 3. Indeed, D-opt can only ensure the uniqueness of predecessor for each observation but not the uniqueness of successor. In contrast, our algorithm enforces the uniqueness constraints of both predecessor and successor explicitly, thus prevents a single trajectory from bifurcating. Moreover, in D-opt the disagreements among solutions of different subproblems is resolved by using a somewhat heuristic voting scheme. On the contrary, in our method subproblems reach consensus in a more rigorous way, that is, by adjusting the parameters $\theta$ of each subproblem iteratively according to the subgradient optimization principle. Particularly, in L-DD subproblems are guaranteed to reach agreement on their solutions.

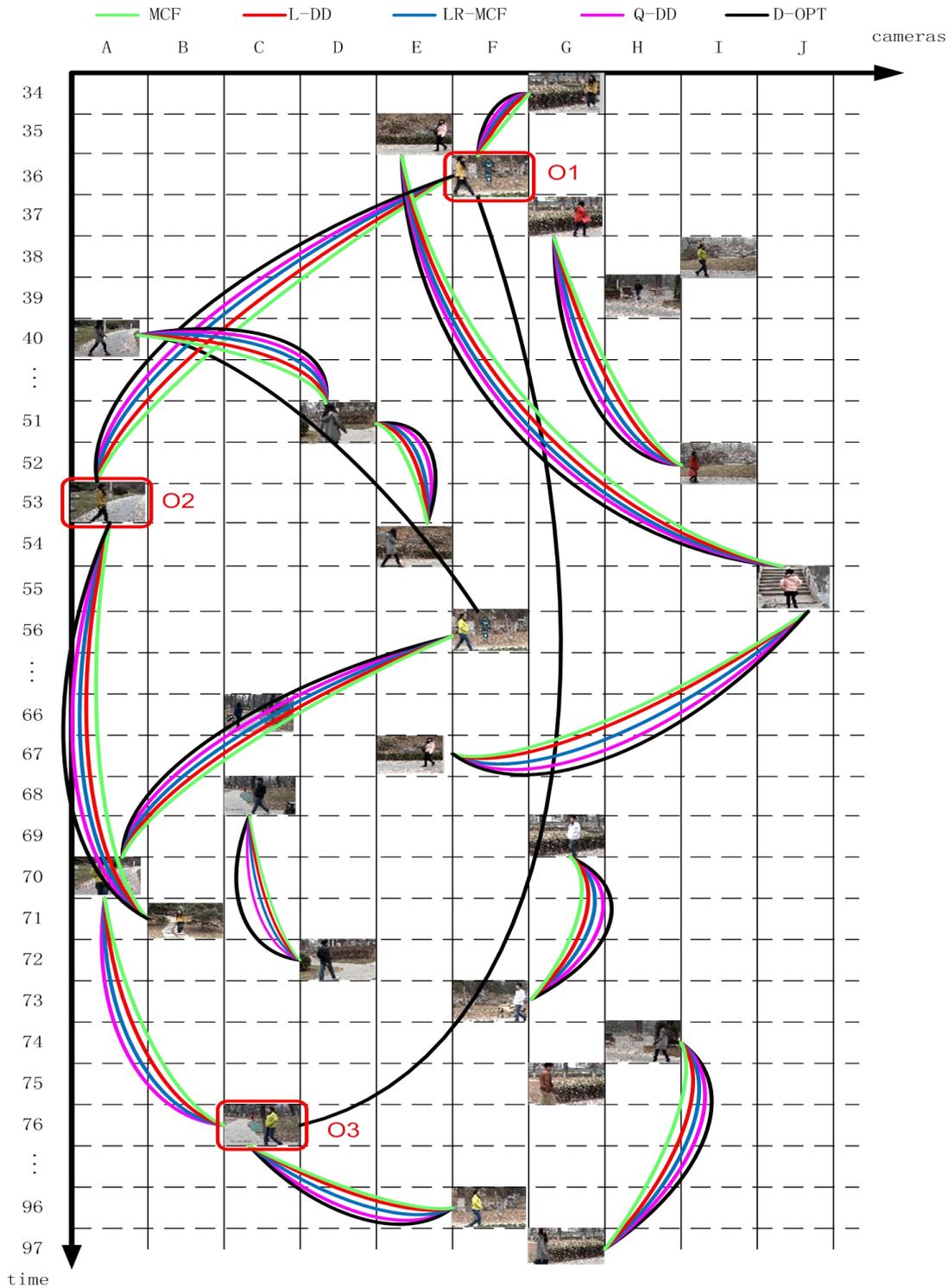

**Fig.10 Linking results. Column corresponds to camera site, and row corresponds to time instance. The linked observations represent trajectory of a single person. We use different colors to represent linking results of different algorithms. Note that observation O1 has two successors, O2 and O3, according to D-opt algorithm.**

*G. Q-DD vs. LR-MCF*

We apply the LR-MCF algorithm proposed in [23] to our problem, which exploits the higher order constraints for improving data association. LR-MCF works in an iterative manner. In each iteration, we solve a MCF problem (41) using the algorithm in [22], and update parameter $\lambda$ according to Eq.(42) for the next iteration. The initial value of $\lambda$ is chosen as 0. The cost-flow network underlying problem (41) is shown as in Fig.11. The numbers of nodes and edges of the network in our problems are extremely large. The underlying networks include about $1.8 \times 10^4$ nodes and $7 \times 10^5$ edges in the two experiments. Therefore, solving MCF problem in each iteration is very time consuming. In our implementation, it takes about an average of 22 seconds to solve the MCF problem of such scale even by using a highly efficient algorithm [22].

The optimal value of problem (41), or the dual energy, provides a lower bound of problem (38), which is plotted in Fig.7. In principle, we can find the solution of problem (38) at each iteration, by removing the violations of uniqueness constraint following the approach in [23], and compute the primal energy. However, in practice, we find that the number of possible "competing tracks" grows exponentially with the number of violations. At the beginning of the algorithm, the number of violations is large. We cannot obtain the primal energy with reasonable computational effort, so we do not plot the primal energy. It is clear from Fig.7 that after 879 iterations the lower bound reaches the optimal value of 4081 in Campus Garden experiment and after 1480 iterations the lower bound reaches the optimal value of 6241 in Office Building experiment. In both experiments, the optimal lower bounds obtained by Q-DD and LR-MCF are the same. And from Table 2 and 3 we can see that Q-DD and LR-MCF achieve the same association accuracy. However, LR-MCF is a centralized algorithm which requires all of the data to be transmitted to and processed on a central server, while Q-DD is a distributed algorithm where the information is processed on each smart camera. It is also interesting to compare the convergence behavior of the two methods. From Fig.7 (a) and (b), it appears that the lower bound of LR-MCF approach the optimal value rapidly at the first tens of iterations. But we can see from Fig.7 (c) and (d), which are transformed from Fig.7 (a) and (b) by rescaling the horizontal axis into real time, that for any given time of computation, Q-DD always provides a better lower bound than LR-MCF, and generally, a better data association accuracy.

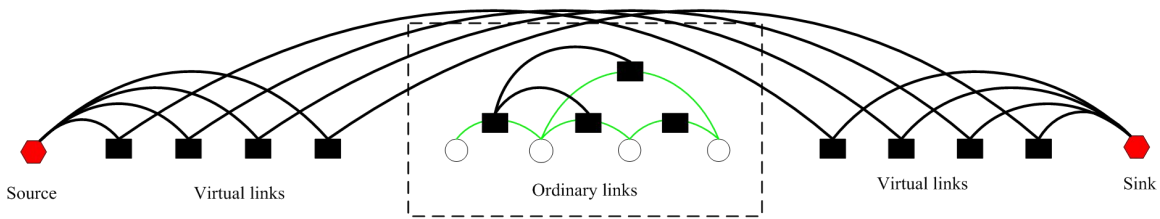

**Fig.11 Cost-flow network underlying LR-MCF. The white circles represent observations. Black rectangles represent links between observations (including virtual link), which are the nodes of the cost-flow network. For a given observation, each pair of its incoming and outgoing link nodes are connected by an edge in the cost-flow network. The red hexagons represent source and sink nodes, which are connected with virtual link nodes only. We do not plot the edges between virtual and ordinary links to avoid clutter. The incoming virtual link of observation $y_i$ should be connected with every outgoing links of $y_i$, and the outgoing virtual link of observation $y_i$ should be connected with every incoming links of $y_i$.**

5. Conclusion

In this paper, we describe a new approach for distributed data association in smart camera networks. The problem is formulated as finding the optimal linking configuration of observations generated by the networks, and the resulting IP problem is solved using dual decomposition technique. Specifically, we propose two distributed algorithms, L-DD and Q-DD, both of which are very simple but powerful. We present the theoretical analysis of the property of the two algorithms and show their superiority through extensive comparisons with state-of-the-art methods on two real datasets.

In our view, there are several interesting directions deserving further investigation. First, in our implementation, the appearance similarity is calculated as the Bhattacharya distance between normalized histogram of pixels. But this simple feature and distance measure may be not effective in more challenging scenario, where appearance of the same person may change significantly due to variations in view angle and lighting condition. We note that lots of powerful appearance descriptors and distance learning methodologies have been proposed in person re-identification research community [2-6]. Our framework is flexible in that any of these techniques can be incorporated into our framework directly as long as they can provide a similarity measure between a pair of appearance observations. Second, in our algorithms, the dual problem is solved by using projected subgradient algorithm. The main limitation of subgradient based method is that the convergence can be slow especially when there are a large number of overlapping variables between subproblems. We note that recently a new algorithm for MAP inference in graphical model called AD$^3$ was proposed [42], which ally the simplicity of dual decomposition and the effectiveness of augmented Lagrangian methods, resulting in a significant speeding up of the convergence rate, namely, reducing iterations bound from $O(1/\varepsilon^2)$ to $O(1/\varepsilon)$. We hope we can exploit this technique in our problem in the future. Finally, we plan to extend the use of our method to more realistic scenarios, where the size of networks is larger, the duration of video is longer and the camera scene is more crowded. We are working in these directions.

Acknowledgement

This work is supported by the National Natural Science Foundation of China, under grant No.61174020.